\documentclass{article} 
\usepackage{nips15submit_e,times}
\usepackage{times}
\usepackage[fleqn]{amsmath}
\usepackage{amssymb}

\usepackage{graphicx} 
\usepackage{subfigure} 

\usepackage{tikz}
\usetikzlibrary{bayesnet}
\usepackage{url}

\usepackage[numbers]{natbib}

\usepackage{afterpage}
\usepackage{float}
\usepackage{hyperref}

\usepackage{algorithm}
\usepackage{algorithmic}

\usepackage{booktabs}

\floatname{algorithm}{Algorithm}

\title{CRAFT: ClusteR-specific Assorted Feature selecTion}
\begin{document}

\author{
Vikas K.~Garg \\ 
Computer Science and Artificial Intelligence Laboratory (CSAIL)\\
Massachusetts Institute of Technology (MIT)\\
\texttt{vgarg@csail.mit.edu} \\
\And
Cynthia Rudin \\
Computer Science and Artificial Intelligence Laboratory (CSAIL)\\
Massachusetts Institute of Technology (MIT)\\
\texttt{rudin@mit.edu} \\
\And
Tommi Jaakkola \\
Computer Science and Artificial Intelligence Laboratory (CSAIL)\\
Massachusetts Institute of Technology (MIT)\\
\texttt{tommi@csail.mit.edu} \\
}
\maketitle

\begin{abstract} 
We present a framework for clustering with cluster-specific feature selection. The framework, CRAFT, is derived from asymptotic log posterior formulations of nonparametric MAP-based clustering models. CRAFT handles assorted data, i.e., both numeric and categorical data, and the underlying objective functions are intuitively appealing. The resulting algorithm is simple to implement and scales nicely, requires minimal parameter tuning, obviates the need to specify the number of clusters a priori, and compares favorably with other methods on real datasets.
\end{abstract} 




\section{Introduction}
%
    
We present a principled framework for clustering with feature selection. Feature selection can be global (where the same features are used across clusters) or local (cluster-specific). For most real applications, feature selection ideally should be cluster-specific, e.g., when clustering news articles, the similarity between articles about politics should be assessed based on the language about politics, regardless of their references to other topics such as sports. However, choosing cluster-specific features in an unsupervised way can be challenging. In fact, unsupervised \textit{global} feature selection is widely considered a hard problem \cite{NDFS}. \textit{Cluster-specific} unsupervised feature selection is even harder since separate, possibly overlapping, subspaces need to be inferred. Our proposed method, called CRAFT (ClusteR-specific Assorted Feature selecTion), has a prior parameter that can be adjusted for a desired balance between global and local feature selection. 

CRAFT addresses another challenge for clustering: handling assorted data, containing both numeric and categorical features. The vast majority of clustering methods, like K-means \cite{Lloyd,Macqueen}, were designed for numeric data. However, most real datasets contain categorical variables or are processed to contain categorical variables; for instance, in web-based clustering applications, it is standard to represent each webpage as a binary (categorical) feature vector. Treating categorical data as if it were real-valued does not generally work since it ignores ordinal relationships among the categorical labels. This explains why despite several attempts (see, e.g., \cite{Ahmad,Aranganayagi,Huang1,Huang2,San}), variations of K-means have largely proved ineffective in handling mixed data.

The derivations of CRAFT's algorithms follow from asymptotics on the log posterior of its generative model. The model is based on Dirichlet process mixtures \cite{Guan, Shaf, Sohn} (see \citet{Kim14} for a prototype model with feature selection), and thus the number of clusters can be chosen non-parametrically by the algorithm. Our asymptotic calculations were inspired by the works of \citet{Kulis}, who derived the DP-means objective by considering approximations to the log-likelihood, and \citet{Broderick}, who instead approximated the posterior log likelihood to derive other nonparametric variations of K-means. These works do not consider feature selection, and as a result, our generative model is entirely different, and the calculations differ considerably from previous works.  However, when the data are only numeric, we recover the DP-means objective with an additional term arising due to feature selection. CRAFT's asymptotics yield interpretable objective functions, and suggest K-means-style algorithms that recovered subspaces on synthetic data, and outperformed several state-of-the-art benchmarks on real datasets in our experiments. 





\section{The CRAFT Framework}\label{SectionCRAFTWithFS}
The main intuition behind our formalism is that the points in a cluster should agree closely on the features selected for that cluster. As it turns out, the objective is closely related to the cluster's entropy for discrete data and variance for numeric data. For instance, consider a parametric setting where the features are all binary categorical, taking values only in $\{0, 1\}$, and we select all the features. Assume that the features are drawn from independent Bernoulli distributions.  Let the cluster assignment vector be $z$, i.e., $z_{n, k} = 1$ if point $x_n$ is assigned to cluster $k$. Then,  we obtain the following objective using a straightforward maximum likelihood estimation (MLE) procedure:    
\small \begin{eqnarray*}\label{BinaryClusterObjective2}
\arg\!\min_{z}\,\,  \sum_{k}^{} \sum_{n: z_{n, k} = 1}  \sum_{d} \mathbb{H}(\mu_{kd}^*)
\end{eqnarray*} \normalsize
where $\mu_{kd}^*$ denotes the mean of feature $d$ computed by using points belonging to cluster $k$, and the entropy function $\mathbb{H}(p)  =  -p \log p - (1 - p) \log (1 - p)\, \text{ for } p \in [0, 1]$ characterizes the uncertainty. Thus the objective tries to minimize the overall uncertainty across clusters and thus forces similar points to be close together, which makes sense from a clustering perspective.  

It is not immediately clear how to extend this insight about clustering to cluster-specific feature selection. CRAFT combines assorted data by enforcing a common Bernoulli prior that selects features, regardless of whether they are categorical or numerical. We derive an asymptotic approximation for the posterior joint log likelihood of the observed data, cluster indicators, cluster means, and feature means. Modeling assumptions are then made for categorical and numerical data separately; this is why CRAFT can handle multiple data types. 
Unlike generic procedures, such as Variational Bayes, that are typically computationally intensive, the CRAFT asymptotics lead to elegant K-means style algorithms that have following steps repeated: (a) compute the ``distances"  to the cluster centers using the selected features for each cluster, choose which cluster each point should be assigned (and create new clusters if needed), and recompute the centers and select the appropriate cluster-specific features for the next iteration. 

Formally, the data $x$ consists of $N$ i.i.d. D-dimensional binary vectors $x_1, x_2, \ldots, x_N$.  We assume a Dirichlet process (DP) mixture model  to avoid having to specify a priori the number of clusters $K^{+}$, and use the hyper-parameter $\theta$, in the underlying exchangeable probability partition function (EFPF) \cite{Pitman}, to tune the probability of starting a new cluster.   We use $z$ to denote cluster indicators:  $z_{n, k} = 1$ if $x_n$ is assigned to cluster $k$. Since $K^+$ depends on $z$, we will often make the connection explicit by writing $K^+(z)$. Let $Cat$ and $Num$ denote respectively the set of categorical and the set of numeric features respectively.

The variables $v_{kd} \in \{0, 1\}$ indicate whether feature $d \in [D]$ is selected in cluster $k \in [K]$. 
We assume $v_{kd}$ is generated from a Bernoulli distribution with parameter $\nu_{kd}$. 
Further, we assume $\nu_{kd}$ is generated from a Beta prior having variance $\rho$ and mean $m$.

For categorical features, the features $d$ selected in any cluster $k$ have values drawn from a discrete distribution with parameters $\eta_{kdt},\, d \in Cat$, where $t \in \mathcal{T}_{d}$ indexes the different values taken by the categorical feature $d$. The parameters $\eta_{kdt}$ are drawn from a Beta distribution with parameters $\alpha_{kdt}/K^+$ and 1. 
On the other hand, we assume the values for features that have not been selected are drawn from a discrete distribution with cluster-independent mean parameters $\eta_{0dt}$.  

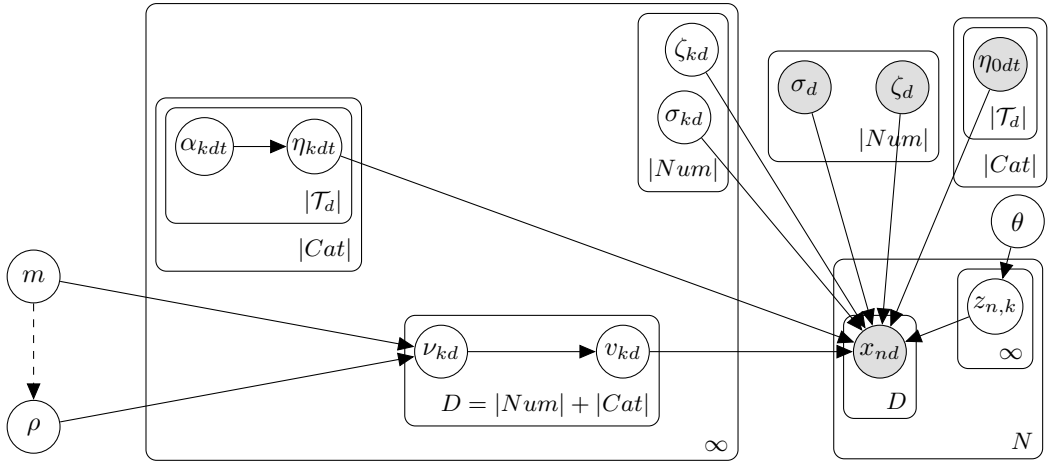
\begin{figure*}[t!]
\centering
\begin{tikzpicture}[x=1.7cm,y=0.8cm] 


  \node[obs]                   (X)      {$x_{nd}$} ; %
  \node[latent, left=of X, xshift=-1cm ]    (Y)      {$v_{kd}$} ; %
  \node[latent, left=of Y]    (nu)      {$\nu_{kd}$} ; %
  \node[latent, left=of nu, xshift= -3.0cm, yshift=1cm]    (M)      {$m$} ; %
   \node[latent, left=of nu,  xshift= -3.0cm, yshift=-1cm]    (rho)      {$\rho$} ;
   \node[latent, above=of Y, xshift=-4.1cm, yshift=1.2cm]    (eta)      {$\eta_{kdt}$} ;
   \node[obs, above=of X, yshift= 2cm, xshift=0.3cm] (meanGlobal) {$\zeta_{d}$};
   \node[obs, above=of X, yshift= 2cm, xshift=-1cm] (sigmaGlobal) {$\sigma_{d}$};
   \node[latent, above=of X, xshift=-2.6cm, yshift=1.6cm]    (sigma) {$\sigma_{kd}$};
   \node[latent, above=of X, xshift=-2.5cm, yshift=2.5cm ]    (zeta)      {$\zeta_{kd}$} ;
   \node[obs, above=of X, yshift=2.3cm, xshift=1.6cm]    (etad) {$\eta_{0dt}$};
   \node[latent, left=of eta, xshift=1cm] (alpha) {$\alpha_{kdt}$};
   \node[latent, right=of X, yshift=0.6cm, xshift=-0.9cm ]    (Z)      {$z_{n,k}$} ;
   \node[latent, above=of Z, xshift=0.3cm, yshift=-0.4cm]    (theta)      {$\theta$} ; 
   \edge[dashed, ->] {M}{rho};

\edge{M}{nu}
\edge{meanGlobal}{X}
\edge{rho}{nu}
\edge{nu}{Y}
\edge{eta}{X}
\edge{etad}{X}
\edge{sigmaGlobal}{X}
 \edge {Y} {X}
 \edge{zeta}{X}
 \edge{Z}{X}
 \edge{sigma}{X}
 \edge{theta}{Z}
 \edge{alpha}{eta}
 \plate{etaT}{(etad)}{$\vert \mathcal{T}_d \vert$}
 \plate{etaTD}{(etaT)}{$|Cat|$}
 \plate{}{(meanGlobal)(sigmaGlobal)}{$\vert Num \vert$}
 \plate{plateNY}{(nu)(Y)}{$D = |Num| + |Cat|$}
 \plate{plateEta}{(eta)(alpha)}{$\vert \mathcal{T}_d \vert$}
 \plate{plateCat}{(plateEta)}{$\vert Cat \vert$}
 \plate{plateZeta}{(zeta)(sigma)}{$\vert Num \vert$}
 \plate{}{(plateNY)(plateCat)(plateZeta)}{$\infty$}
 \plate{plateX}{(X)}{$D$}
 \plate{plateZ}{(Z)}{$\infty$}
\plate{plateZX}{(plateZ)(plateX)}{$N$}
 

 \end{tikzpicture}
 




 \caption{CRAFT- Graphical model. For cluster-specific feature selection $\rho$ is set to a high value determined by $m$, whereas for global feature selection $\rho$ is set close to 0. The dashed arrow emphasizes this important part of our formalism that unifies cluster-specific and global feature selection. \label{fig:FDPCRAFT_mix-Graphical}}
\end{figure*}

For numeric features, we formalize the intuition that the features selected to represent clusters should exhibit small variance relative to unselected features by assuming a conditional density of the form:
$$f(x_{nd}|v_{kd}) = \dfrac{1}{Z_{kd}} e^{-\left[v_{kd} \dfrac{(x_{nd}-\zeta_{kd})^2}{2 \sigma_{kd}^2} + (1-v_{kd}) \dfrac{(x_{nd}-\zeta_{d})^2}{2 \sigma_d^2}\right]}, \,\,\,\,  Z_{kd} = \dfrac{\sqrt{2\pi} \sigma_d \sigma_{kd}}{\sigma_{kd} \sqrt{1-v_{kd}} + \sigma_d \sqrt{v_{kd}}},$$
where $x_{nd} \in \mathbb{R}$, $v_{kd} \in \{0, 1\}$, and $Z_{kd}$ ensures $f$ integrates to 1, and $\sigma_{kd}$ guides the allowed variance of a selected feature $d$ over points in cluster $k$ by asserting feature $d$ concentrate around its cluster mean $\zeta_{kd}$. The features not selected are assumed to be drawn from Gaussian distributions that have cluster independent means $\zeta_d$ and variances $\sigma_{d}^2$. Fig. \ref{fig:FDPCRAFT_mix-Graphical} shows the graphical model.   

Let $\mathbb{I}(\mathcal{P})$ be 1 if the predicate $\mathcal{P}$ is true, and 0 otherwise. Under asymptotic conditions, minimizing the joint negative log-likelihood yields the following objective (see the Supplementary for details):  

\small
\begin{eqnarray}\nonumber \label{Discrepancy} 
\arg\!\!\!\min_{z, v, \eta, \zeta, \sigma}\, \underbrace{\sum_{k=1}^{K^+} \sum_{n: z_{n, k} = 1}  \sum_{d \in Num} \dfrac{v_{kd} (x_{nd} - \zeta_{kd})^2}{2\sigma_{kd}^2}}_{\text{Numeric Data Discrepancy}} + \underbrace{(\lambda + DF_0) K^+}_{\text{Regularization Term}} + \underbrace{\left(\displaystyle \sum_{k=1}^{K^+} \sum_{d= 1}^{D} v_{kd}\right) F_{\Delta}}_{\text{Feature Control}}, \\
+ \underbrace{\sum_{k=1}^{K^+} \sum_{d \in Cat} \left[v_{kd} \left(\sum_{n: z_{n, k} = 1} - \mathbb{I}(x_{nd} = t) \log \eta_{kdt})\right) + (1-v_{kd}) \sum_{n: z_{n, k} = 1} \sum_{t \in \mathcal{T}_d}  - \mathbb{I}(x_{nd} = t) \log \eta_{0dt} \right]}_{\text{Categorical Data Discrepancy}} \nonumber 
\end{eqnarray}    \normalsize
where
$F_{\Delta}$ and $F_0$ depend only on the ($m$, $\rho$) pair: $F_{\Delta}  =  F_1 - F_0$, with
\begin{eqnarray} \label{ComputeF}
F_0 & = & (a_0+b_0) \log\, (a_0+b_0) - a_0 \log \, a_0 - b_0 \log \, b_0,   \nonumber\\
F_1 & = & (a_1+b_1) \log\, (a_1+b_1) - a_1  \log \, a_1 - b_1 \log \, b_1, \\
a_0 & = &  \dfrac{m^2(1-m)}{\rho} - m, b_0  =  \dfrac{m(1-m)^2}{\rho} + m, a_{1}   =  a_0 +  1,  \text{ and }\,\,b_1  =  b_0  -  1. \nonumber
\end{eqnarray}

This objective has an elegant interpretation. The categorical and numerical discrepancy terms show how selected features (with $v_{kd}=1$) are treated differently than unselected features. The
regularization term controls the number of clusters, and  modulates the effect of feature selection. The feature control term contains the adjustable parameters: $m$ controls the number of features that would be turned on per cluster, whereas $\rho$ guides the extent of cluster-specific feature selection. A detailed derivation is provided in the Supplementary.

\begin{algorithm}[t]
\caption{CRAFT}          
\label{alg3}                           
\begin{algorithmic}                    
    \REQUIRE $x_1, \ldots, x_N$: $D$-dimensional input data with categorical features $Cat$ and numeric features $Num$,  $\lambda > 0$: regularization parameter,  and $m \in (0, 1)$: fraction of features per cluster, and (optional) $\rho \in (0, m(1-m))$: control parameter that guides global/local feature selection. Each feature $d \in Cat$ takes values from the set $\mathcal{T}_d$, while each feature $d \in Num$ takes values from $\mathbb{R}$.  
    \ENSURE $K^{+}$: number of clusters, $l_1, \ldots, l_{K^+}$: clustering, and $v_1, \ldots, v_{K^+}$: selected features.
    \begin{enumerate} 
     \STATE Initialize $K^+ = 1$, $l_1 = \{x_1, \ldots, x_N\}$, cluster center (sample randomly) with categorical mean $\eta_1$ and numeric mean $\zeta_1$, and draw $v_1 \sim \left[Bernoulli\left(m \right)\right]^D$.  If $\rho$ is not specified as an input, initialize $\rho = \max\{0.01, m(1-m) - 0.01\}$.  Compute the global categorical mean $\eta_{0}$. 
      Initialize the cluster indicators $z_n = 1$ for all $n \in [N]$, and $\sigma_{1d} = 1$ for all $d \in Num$.
      \STATE Compute $F_{\Delta}$ and $F_0$ using \eqref{ComputeF}.
     \STATE Repeat until cluster assignments do not change
     \begin{itemize}
             \item For each point $x_n$ 
            \begin{itemize}
                \item \hspace*{-0.2 cm} Compute $\forall k \in [K^+]$
              \small  \[\hspace*{-1cm}  d_{nk} = \sum_{d \in Num} v_{kd} \dfrac{(x_{nd} - \zeta_{kd})^2}{2 \sigma_{kd}^2} +  \sum_{d \in Cat: v_{kd} = 0} \sum_{t \in \mathcal{T}_d} - \mathbb{I}(x_{nd} = t) \log \eta_{0dt}  \]\\ \[\hspace*{-0.5cm} +  \displaystyle \left(\sum_{d=1}^Dv_{kd}\right)F_{\Delta}  \,\,+ \displaystyle \sum_{d \in Cat: v_{kd} = 1} \sum_{t \in \mathcal{T}_d} - \mathbb{I}(x_{nd} = t) \log \eta_{kdt}.\] \normalsize
                 \STATE If $\displaystyle \min_k d_{nk} > (\lambda + DF_0)$, set $K^+ = K^+ + 1$, $z_n = K^+$, and draw $$v_{K^+d} \small \sim Bernoulli\left(\dfrac{\sum_{j=1}^{K^+-1} a_{v_{jd}}}{\sum_{j=1}^{K^+-1} (a_{v_{jd}} + b_{v_{jd}})}\right)\,\,\, \forall  d \in [D],$$\normalsize where $a$ and $b$ are as defined in \eqref{ComputeF}. Set $\eta_{K^+}$ and $\zeta_{K^+}$ using $x_n$. Set $\sigma_{K^+d} = 1$ for all $d \in Num$.
                 \STATE Otherwise, set $z_n = \displaystyle \arg\!\min_k d_{nk}$. 
            \end{itemize}  
            \item Generate clusters $l_1, \ldots, l_{K^+}$ based on $z_1, \ldots, z_{K^+}$: $l_k = \{x_n \,|\, z_n = k\}$.
            \item Update the means $\eta$ and $\zeta$, and variances $\sigma^2$, for all clusters.
            \item For each cluster $l_k$, $k \in [K^+]$,  update $v_k$: choose the $m|Num|$ numeric features $d'$  with lowest $\sigma_{kd'}$ in $l_k$, and choose $m|Cat|$ categorical features $d$ with maximum value of $G_{d} - G_{kd}$, where $G_{d} =-\, \sum_{n: z_{n, k} = 1} \sum_{t \in \mathcal{T}_d} \mathbb{I}(x_{nd} = t) \log \eta_{0dt}$ and $G_{kd} = -\,\sum_{n: z_{n, k} = 1} \sum_{t \in \mathcal{T}_d} \mathbb{I}(x_{nd} = t) \log \eta_{kdt}$. \\
               \end{itemize}
    \end{enumerate}
\end{algorithmic}
\end{algorithm}

A K-means style alternating minimization procedure for clustering assorted data, along with feature selection is outlined in Algorithm \ref{alg3}. The algorithm repeats the following steps until convergence: (a) compute the ``distances" to the cluster centers using the selected features for each cluster, (b) choose which cluster each point should be assigned to (and create new clusters if needed), and (c) recompute the cluster centers and select the appropriate features for each cluster using the criteria that follow directly from the model objective and variance asymptotics.  
In particular, the algorithm starts a new cluster if the cost of assigning a point to its closest cluster center exceeds $(\lambda + DF_0)$, the cost it would incur to initialize an additional cluster.  The information available from the already selected features is leveraged to guide the initial selection of features in the new cluster. Finally, the updates on cluster means and feature selection are performed at the end of each iteration.        
 
 \textbf{Approximate Budget Setting for a Variable Number of Features:} 
Algorithm \ref{alg3} selects a fraction $m$ of features per cluster, uniformly across clusters. A slight modification would allow Algorithm \ref{alg3} to have a variable number of features across clusters, as follows: 
specify a tuning parameter $\epsilon_c \in (0, 1)$ and choose all the features $d$ in cluster $k$ for which $G_{d} - G_{kd} > \epsilon_c G_{d}$. Likewise for numeric features, we may simply choose features that have variance less than some positive constant $\epsilon_v$.   As we show later, this slightly modified algorithm recovers the exact subspace on synthetic data in the approximate budget setting for a wide range of $m$. 
 
\section{Discussion}\label{discussionsection}
We discuss special cases and extensions below, which have implications for future work.
\subsubsection*{Recovering DP-means objective on Numeric Data}
CRAFT recovers the DP-means objective \cite{Kulis} in a degenerate setting (see Supplementary):
\small \begin{eqnarray} \arg\!\min_{z}\,\, \sum_{k=1}^{K^+(z)} \sum_{n: z_{n, k} = 1}  \sum_{d} (x_{nd} - \zeta_{kd}^*)^2 +  \lambda K^+(z), \end{eqnarray} \normalsize 
where $\zeta_{kd}^*$ denotes the (numeric) mean of feature $d$ computed by using points belonging to cluster $k$.
 
\subsubsection*{Unifying Global and Local Feature Selection}
An important aspect of CRAFT is that the point estimate of $\nu_{kd}$ is
$$\dfrac{a_{kd}}{a_{kd}+b_{kd}} = \dfrac{\left(\dfrac{m^2(1-m)}{\rho} - m\right) +  v_{kd}}{\dfrac{m(1-m)}{\rho} } = m + \dfrac{(v_{kd} - m)\rho}{m(1-m)} \to \begin{cases}  v_{kd}, \qquad \mbox{ as } \rho \to m(1 - m)\\ m, \qquad \,\,\, \mbox{ as } \rho \to 0.  \end{cases}$$ 
Thus, using a single parameter $\rho$, we can interpolate between cluster specific selection, $\rho \to m(1 - m)$,  and global selection, $\rho \to 0$. Since we are often interested only in one of these two extreme cases, this also implies that we essentially need to specify only $m$, which is often determined by application requirements. Thus, CRAFT requires minimal tuning for most practical purposes.    

\subsubsection*{Accommodating Statistical-Computational Trade-offs} 
We can extend the basic CRAFT model of Fig. \ref{fig:FDPCRAFT_mix-Graphical} to have cluster specific means $m_k$, which may in turn be modulated via Beta priors. The model can also be readily extended to incorporate more informative priors or allow overlapping clusters, e.g., we can do away with the independent distribution assumptions for numeric data, by introducing covariances and taking a suitable prior like the inverse Wishart. The parameters $\alpha$ and $\sigma_d$ do not appear in the CRAFT objective since they vanish due to the asymptotics and the appropriate setting of the hyperparameter $\theta$. Retaining some of these parameters, in the absence of asymptotics, will lead to additional terms in the objective thereby requiring more computational effort. Depending on the available computational resource, one might also like to achieve feature selection with the exact posterior instead of a point estimate. CRAFT's basic framework can gracefully accommodate all such statistical-computational trade-offs.

\section{Experimental Results}\label{SectionExperiments}
We first provide empirical evidence on synthetic data about CRAFT's ability to recover the feature subspaces. We then show how CRAFT outperforms an enhanced version of DP-means that includes feature selection on a real binary dataset. This experiment underscores the significance of having different measures for categorical data and numeric data. Finally, we compare CRAFT with other recently proposed feature selection methods on real world benchmarks. In what follows, the \textit{fixed budget} setting is where the number of features selected per cluster is constant, and the \textit{approximate budget} setting is where the number of features selected per cluster varies over the clusters. We set $\rho = m(1-m) - 0.01$ in all our experiments to facilitate cluster specific feature selection.

\subsubsection*{Exact Subspace Recovery on Synthetic Data}

\begin{figure*}[t!]
\subfigure[Dataset]{
   \includegraphics[trim = 0.6in 2.5in 0.1in 2.5in, clip, scale=0.16]{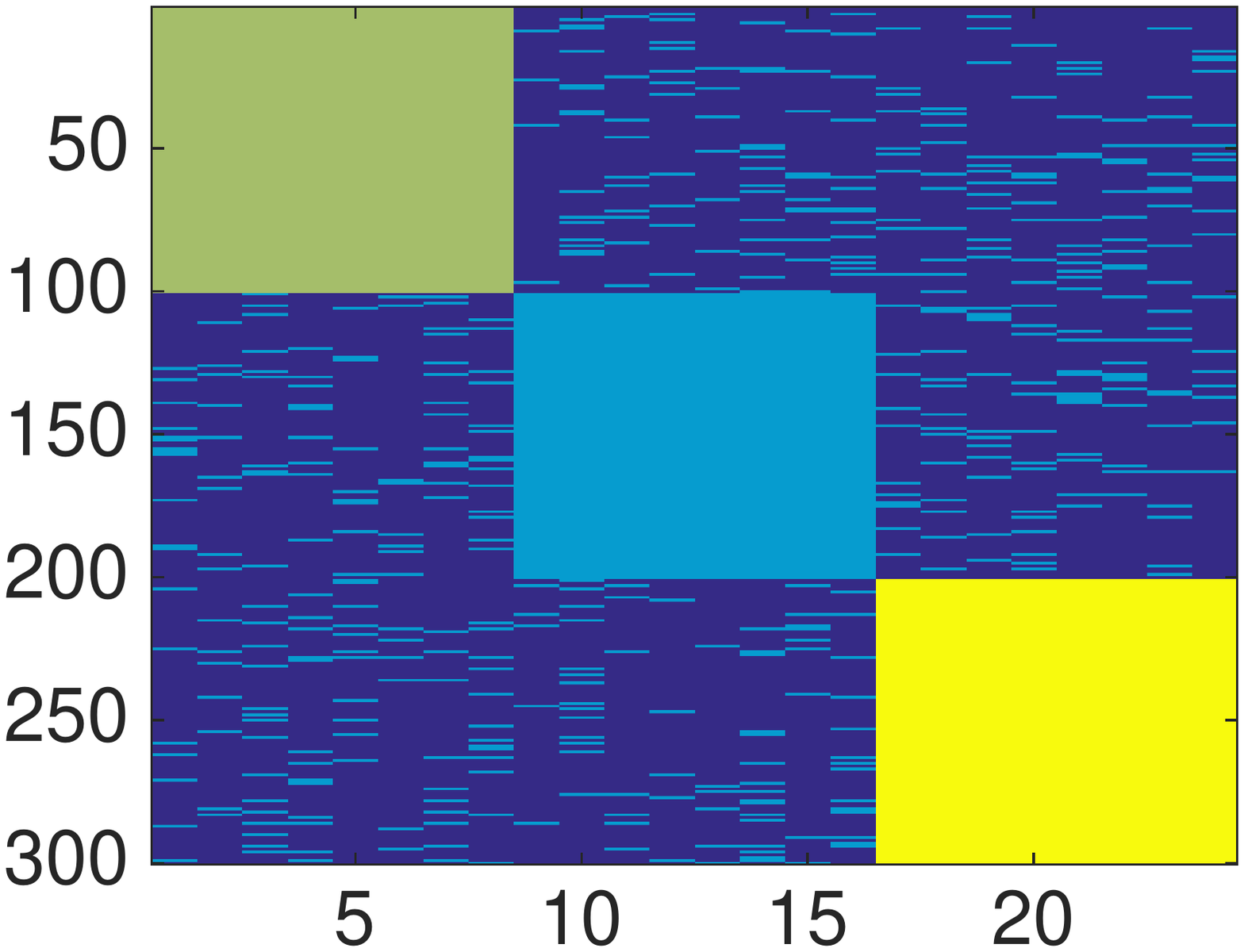}
  \label{subspace1a}
 }
 \subfigure[CRAFT]{
   \includegraphics[trim = 0.6in 2.5in 0.1in 2.5in, clip, scale=0.16]{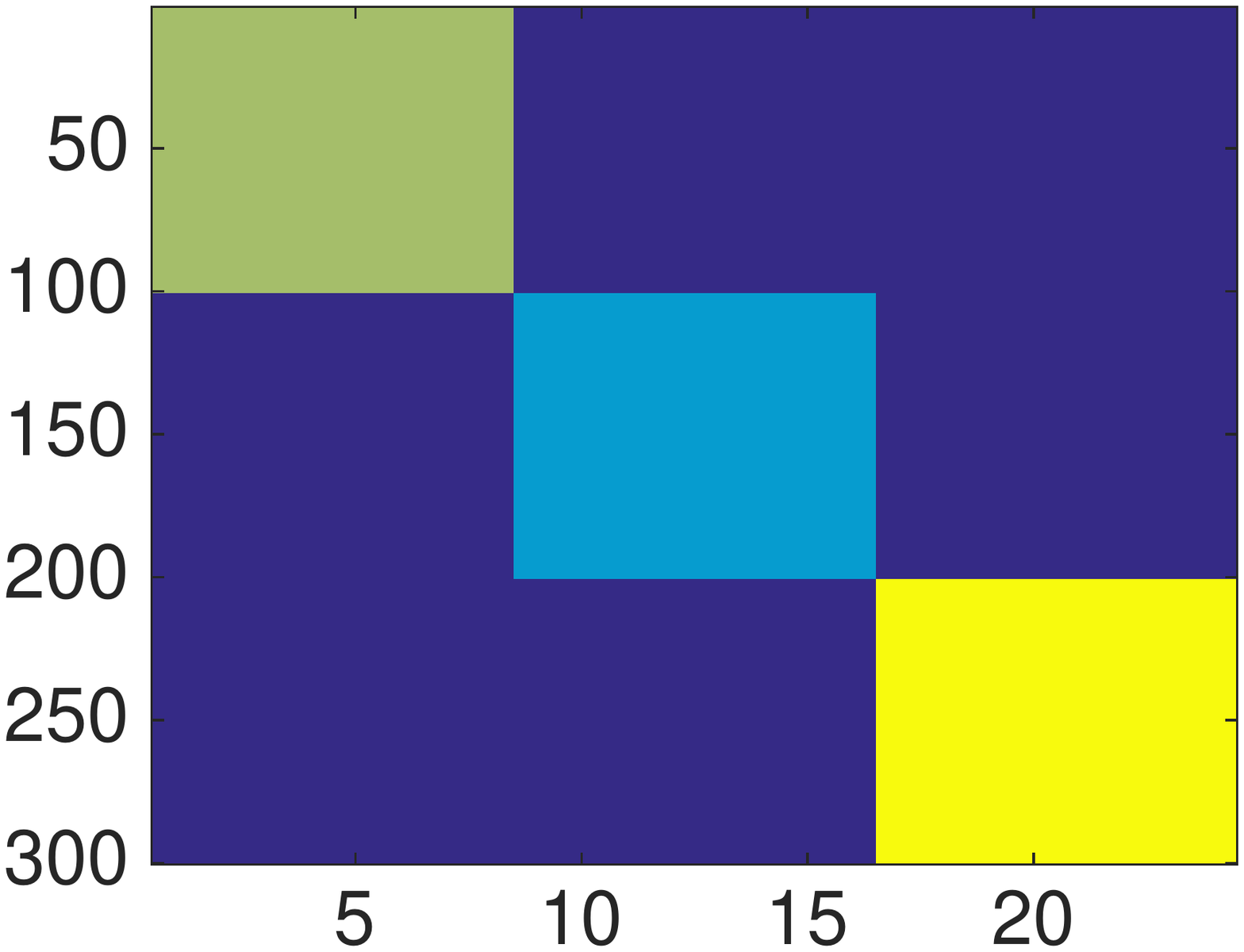}
  \label{subspace1b}  
 }
\subfigure[Dataset]{
   \includegraphics[trim = 0.3in 2.5in 0.1in 2.5in, clip, scale=0.16]{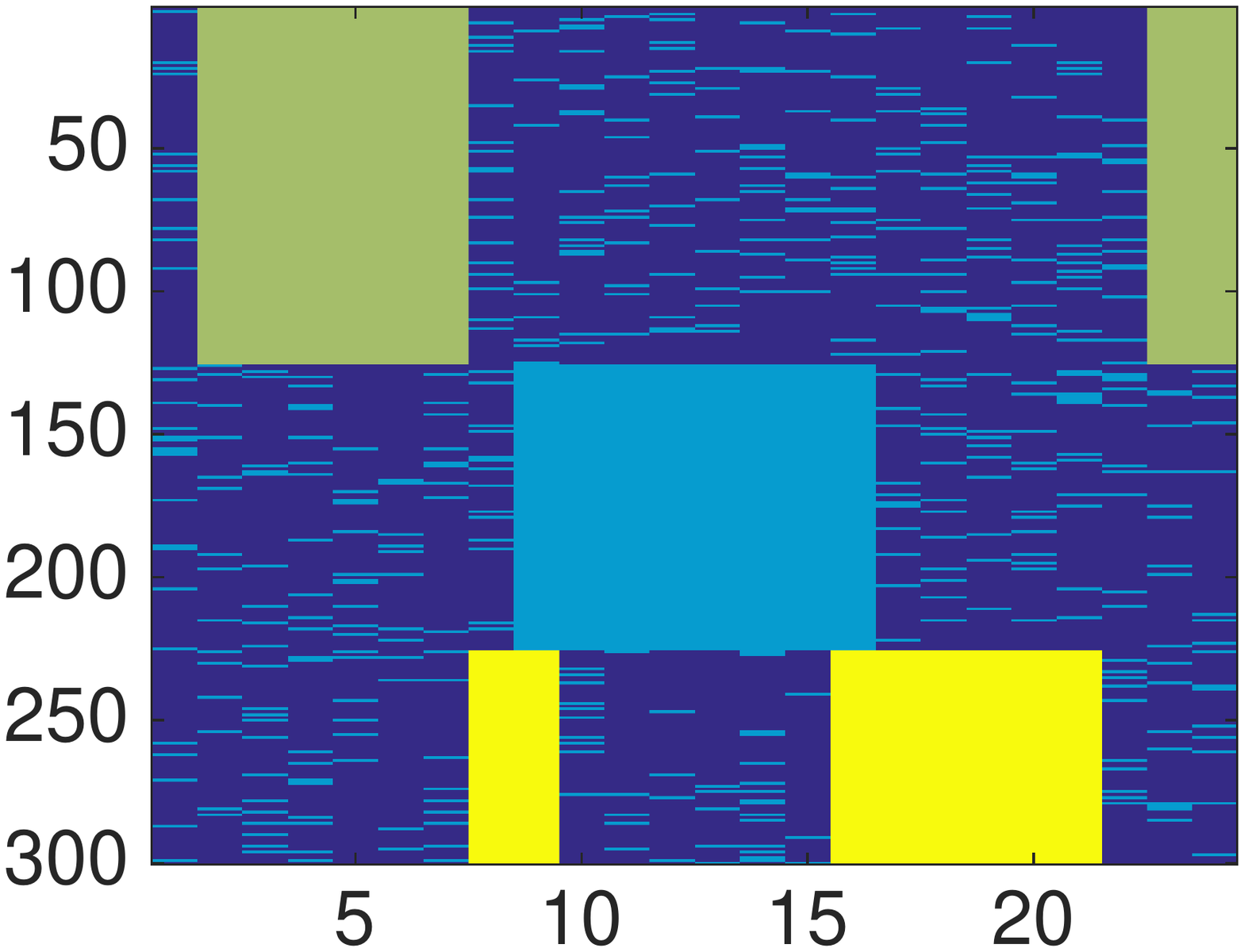}
   \label{subspace1c}
 }
\subfigure[CRAFT]{
   \includegraphics[trim = 0.3in 2.5in 0.1in 2.5in, clip, scale=0.16]{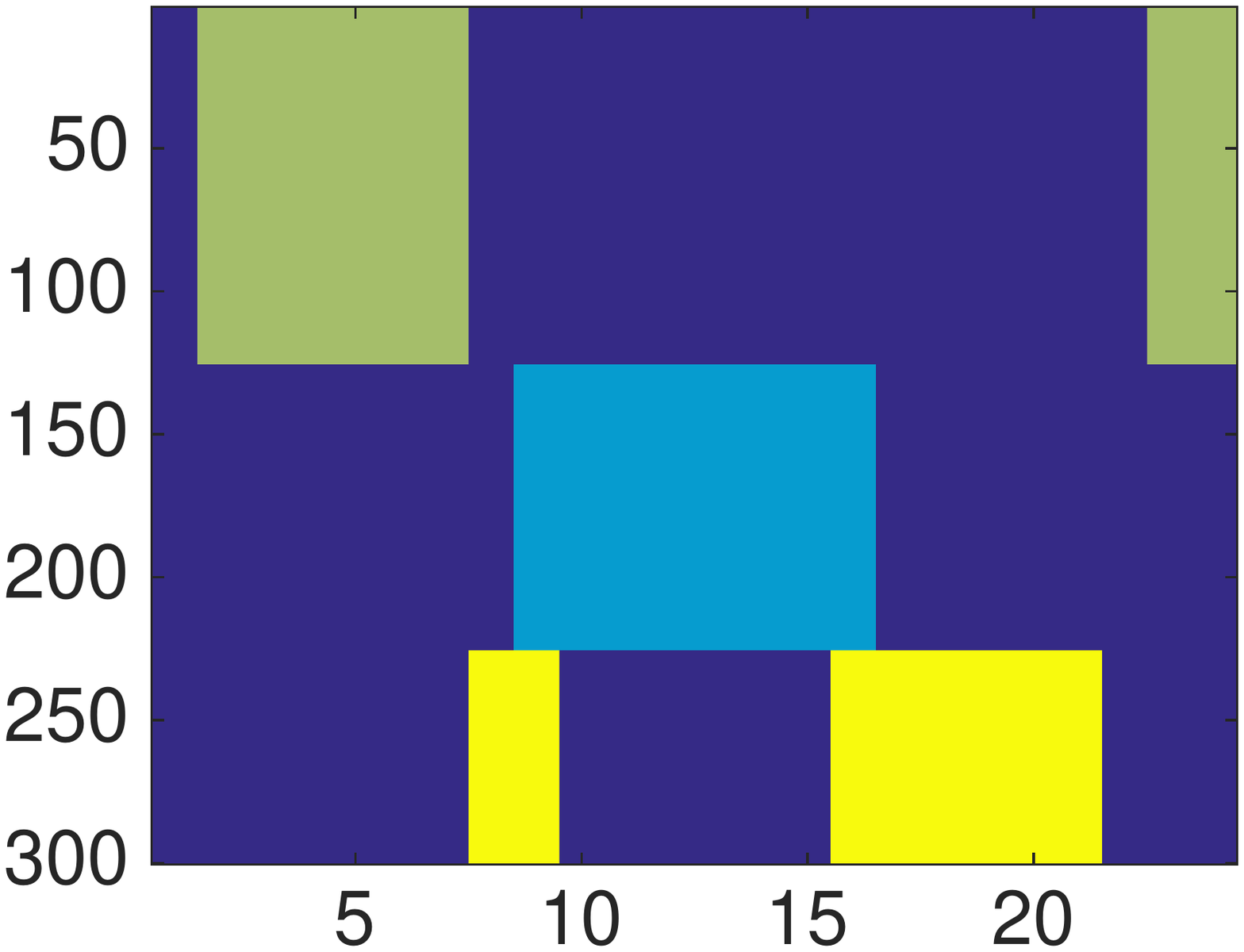}
   \label{subspace1d}
 }
 \label{}
\caption{\textbf{(Fixed budget)} CRAFT recovered the subspaces on categorical datasets.}
\end{figure*}

We now show the results of our experiments on synthetic data, in  both the fixed and the approximate budget settings,  that suggest CRAFT has the ability to recover subspaces on both categorical and numeric data, amidst noise, under different scenarios:  (a) disjoint subspaces, (b) overlapping subspaces including the extreme case of containment of a subspace wholly within the other, (c) extraneous features, and (d) non-uniform distribution of examples and features across clusters.  \\ \\
\noindent \textbf{Fixed Budget Setting:}
Fig.~\ref{subspace1a} shows a binary dataset comprising 300 24-feature points, evenly split between 3 clusters that have disjoint subspaces of 8 features each. We sampled the remaining features independently from a Bernoulli distribution with parameter 0.1. Fig. \ref{subspace1b} shows that CRAFT recovered the subspaces with $m = 1/3$, as we would expect. 
 In Fig. \ref{subspace1c} we modified the dataset to have (a) an unequal number of examples across the different clusters, (b) a fragmented feature space each for clusters 1 and 3, (c) a completely noisy feature, and (d) an overlap between second and third clusters. As shown in Fig. \ref{subspace1d}, CRAFT again identified the subspaces accurately. 

\begin{figure*}[t!]
\centering
\subfigure[Dataset]{
   \includegraphics[trim = 0.9in 2.5in 0.1in 2.8in, clip, scale=0.16]{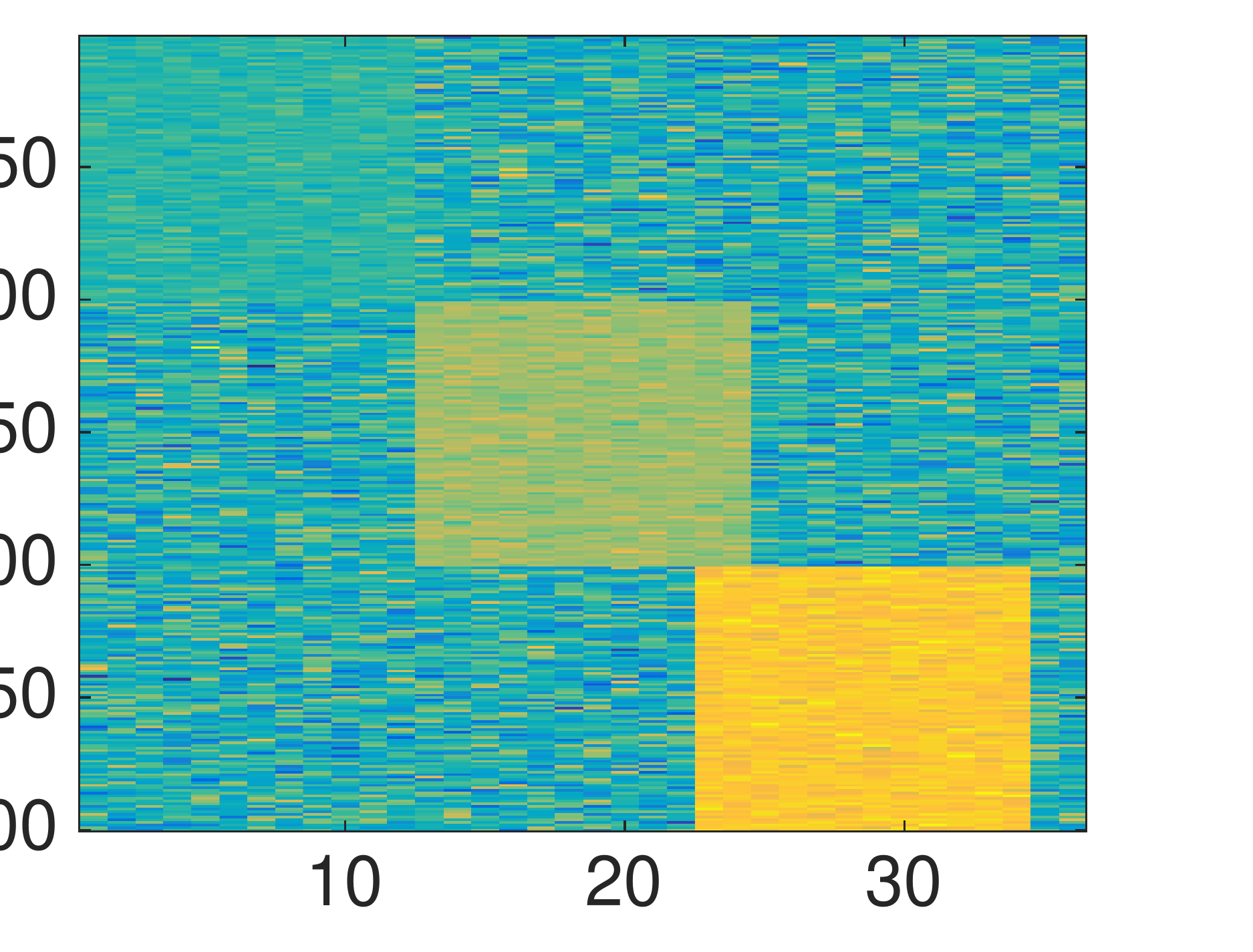}
  \label{subspace2a}
 }
 \subfigure[CRAFT]{
   \includegraphics[trim = 0.9in 2.5in 0.1in 2.8in, clip, scale=0.16]{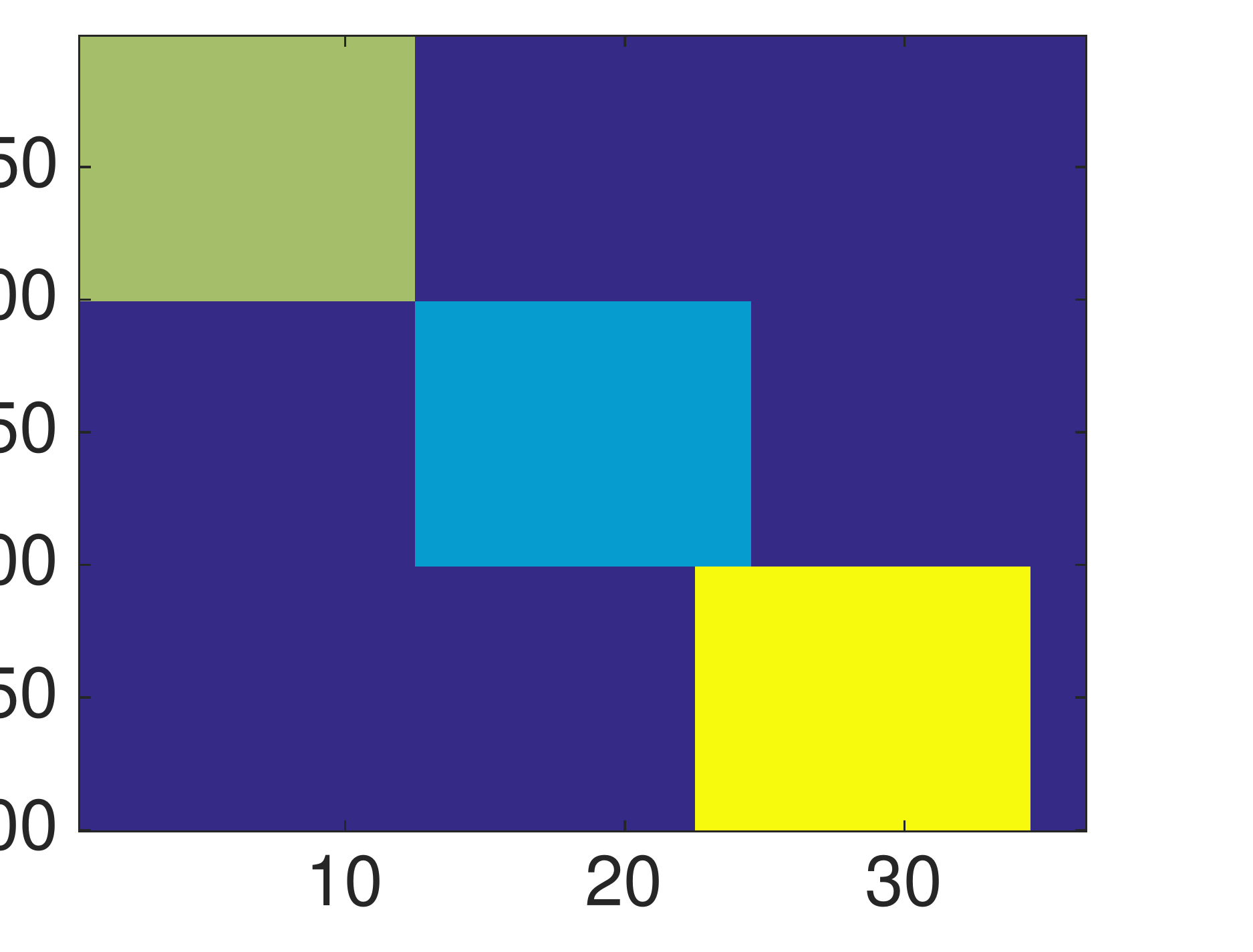}
  \label{subspace2b} 
 }
\subfigure[Dataset]{
   \includegraphics[trim = 0.9in 2.5in 0.1in 2.8in, clip, scale=0.16]{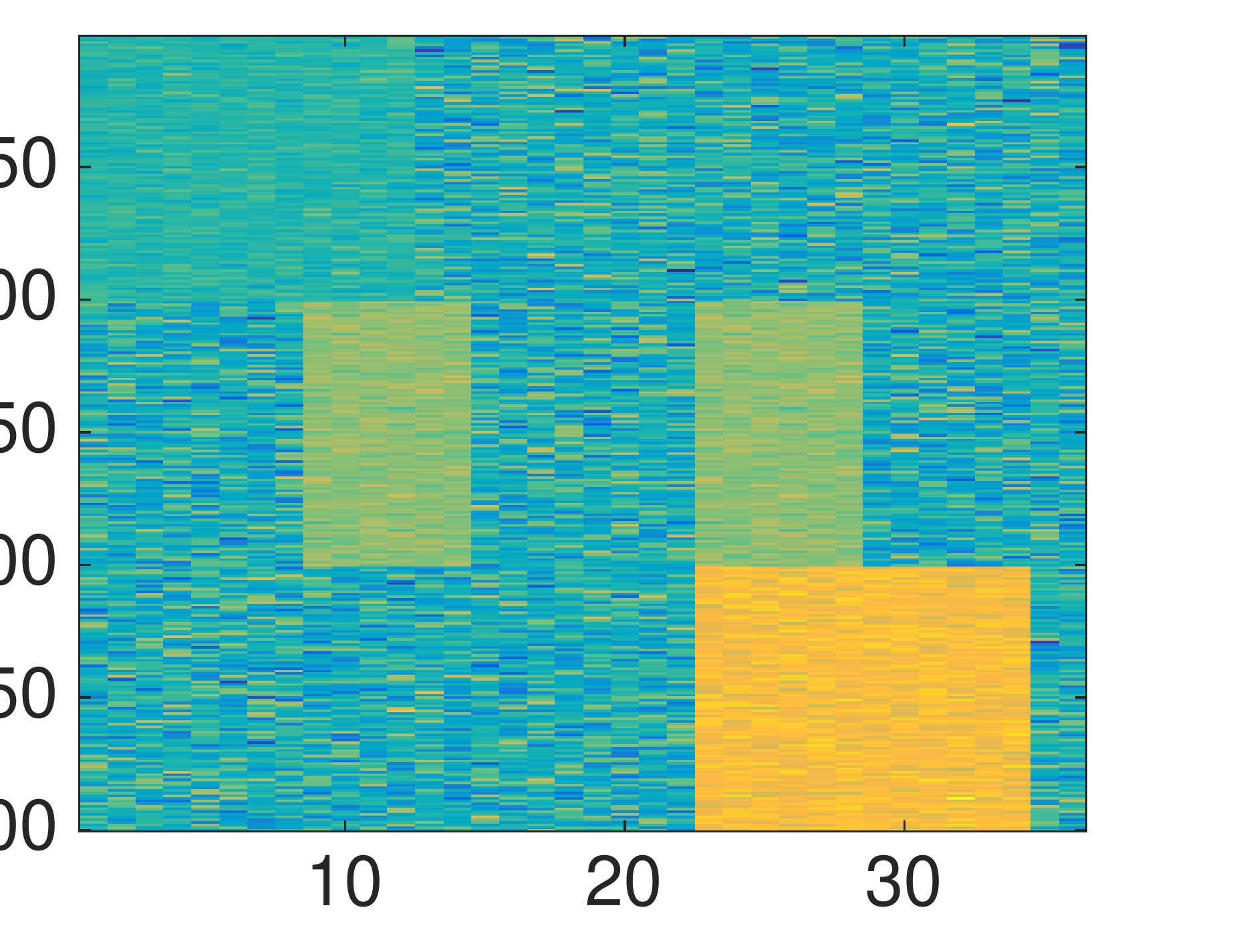}
   \label{subspace2c}
 }
  \subfigure[CRAFT]{
   \includegraphics[trim = 0.9in 2.5in 0.1in 2.8in, clip, scale=0.16]{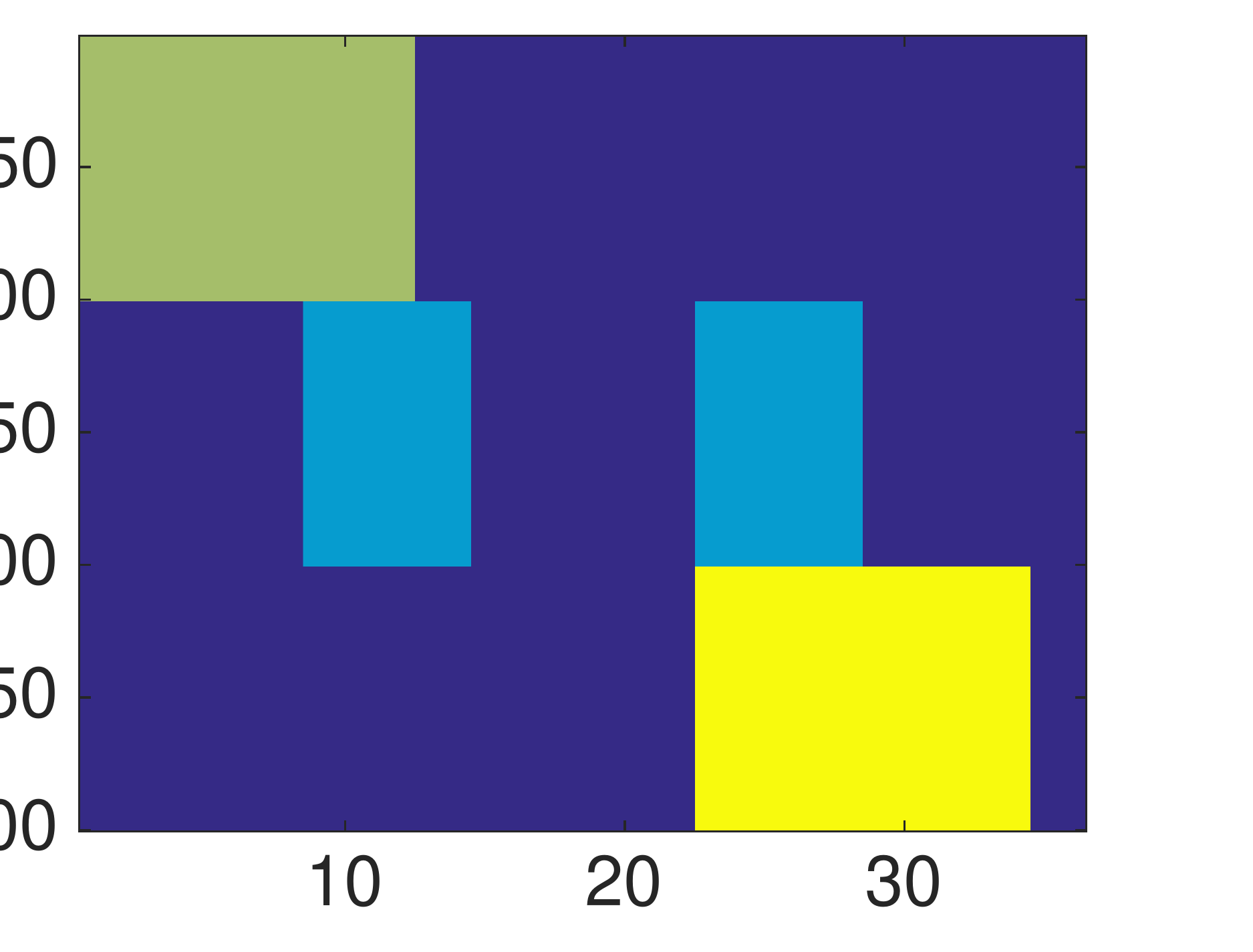}
   \label{subspace2d}
 }
 \label{}
\caption{\textbf{(Fixed budget)} CRAFT recovered the subspaces on numeric datasets.}
\end{figure*}

Fig. \ref{subspace2a} shows the second dataset comprising 300 36-feature points, evenly split across 3 clusters, drawn from independent Gaussians having unit variance and means 1, 5 and 10 respectively. We designed clusters to comprise features 1-12, 13-24, and 22-34 respectively so that the first two clusters were disjoint, whereas the last two some overlapping features.  We added isotropic noise by sampling the remaining features from a Gaussian distribution having mean 0 and standard deviation 3.  Fig. \ref{subspace2b} shows that CRAFT recovered the subspaces with $m = 1/3$. We then modified this dataset in Fig. \ref{subspace2c} to have cluster 2 span a non-contiguous feature subspace. Additionally, cluster 2 is designed such that one partition of its features overlaps partially with cluster 1, while the other is subsumed completely within the subspace of cluster 3. Also, there are several extraneous features not contained within any cluster. CRAFT recovers the subspaces on these data too (Fig. \ref{subspace2d}).   

\noindent \textbf{Approximate Budget Setting:}
We now show that CRAFT may recover the subspaces even when we allow a different number of features to be selected across the different clusters. 

\begin{figure*}[t!]
\centering
\subfigure[Dataset]{
   \includegraphics[trim = 0.6in 2.5in 0.1in 2.5in, clip, scale=0.16]{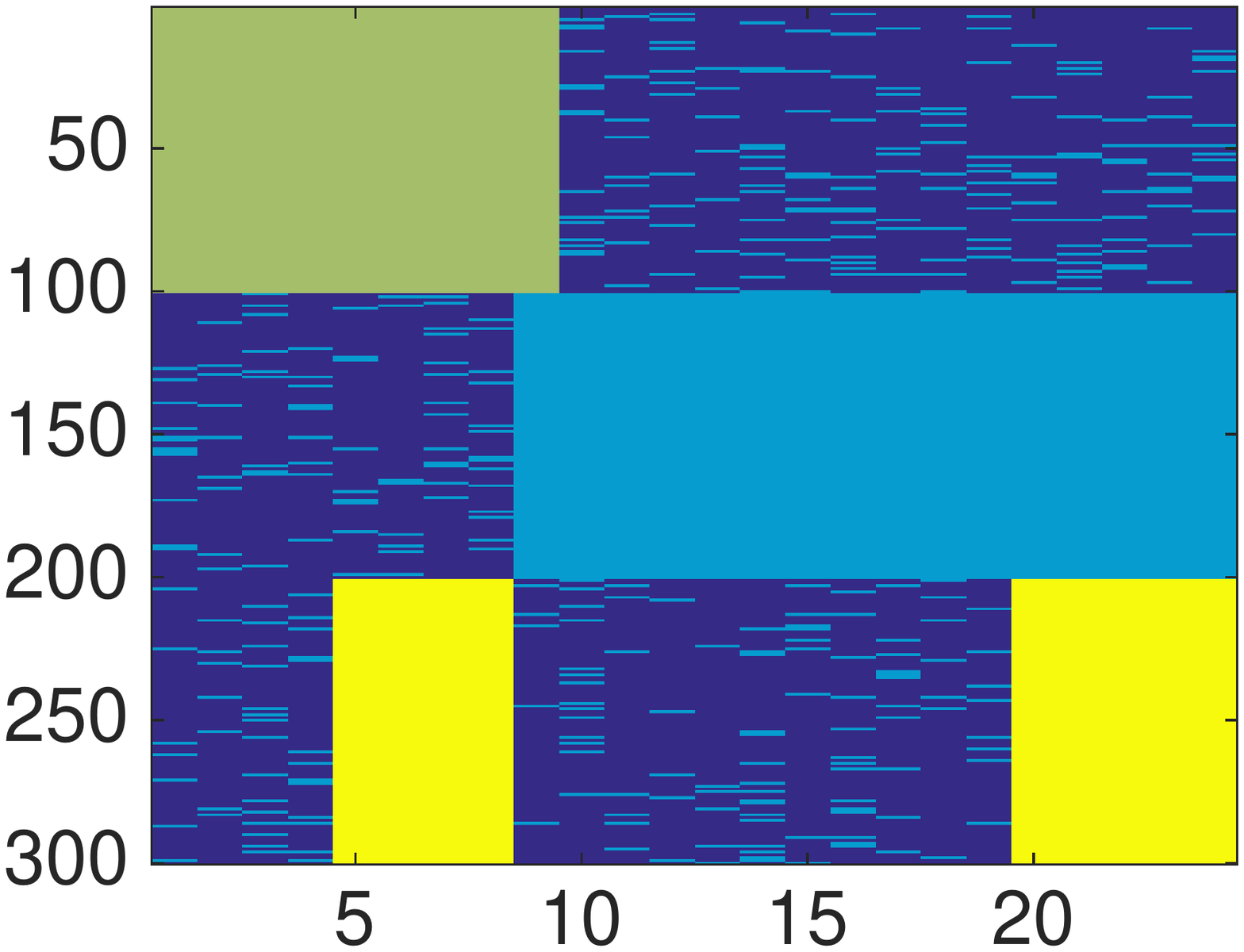}
  \label{subspace3a}
 }
 \subfigure[CRAFT]{
   \includegraphics[trim = 0.6in 2.5in 0.1in 2.5in, clip, scale=0.16]{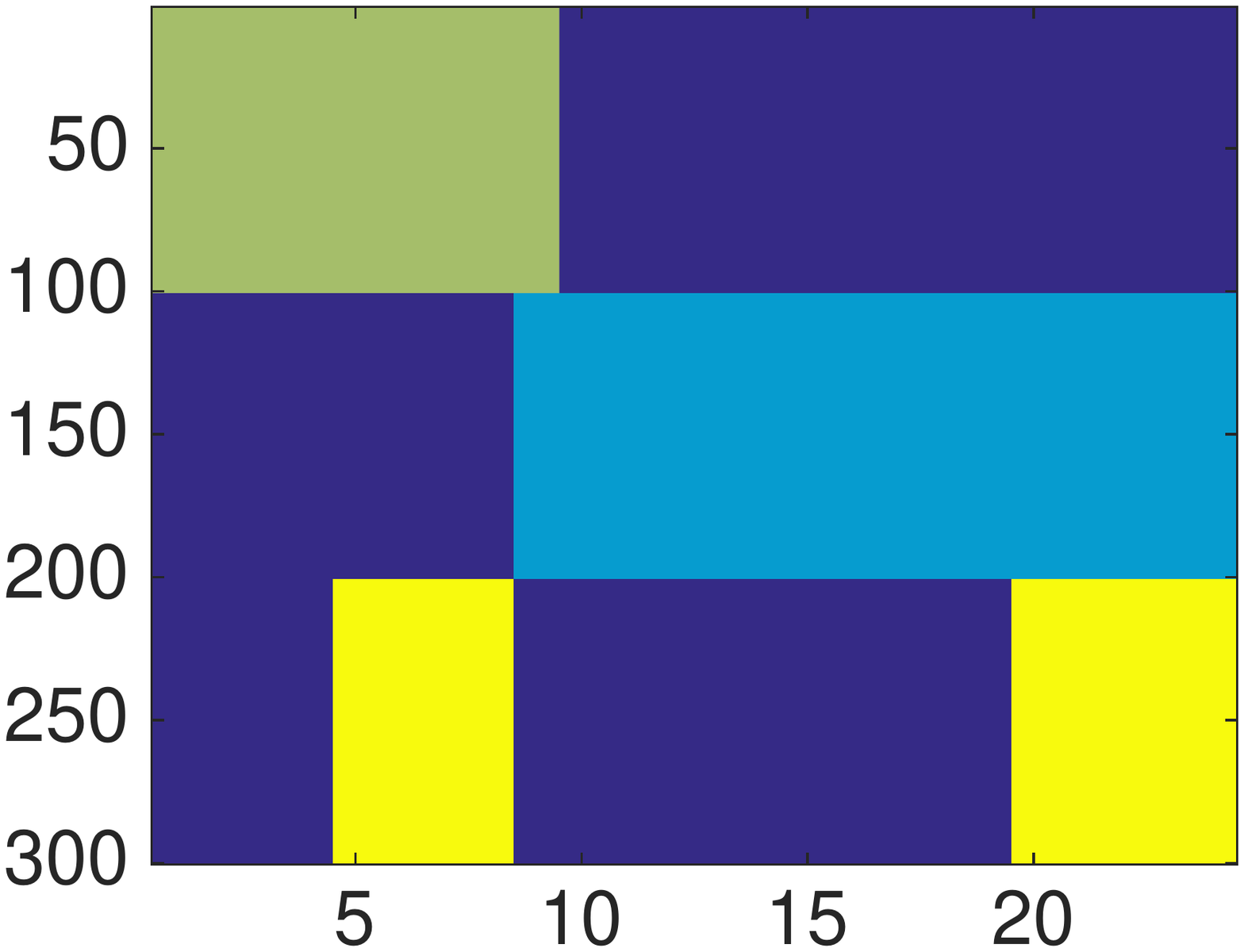}
  \label{subspace3b} 
 }
\subfigure[Dataset]{
   \includegraphics[trim = 0.3in 2.5in 0.1in 2.5in, clip, scale=0.16]{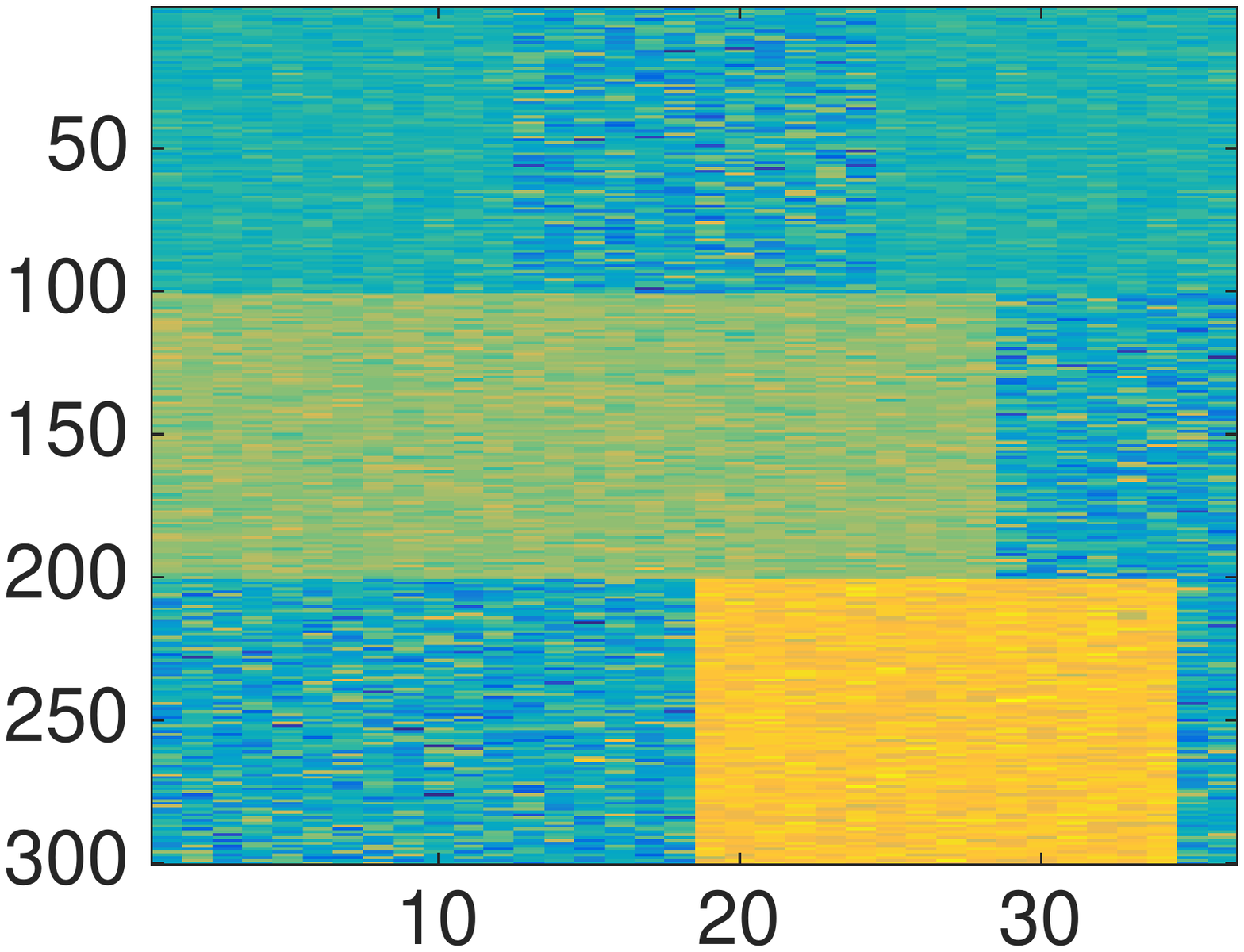}
   \label{subspace3c}
 }
  \subfigure[CRAFT]{
   \includegraphics[trim = 0.3in 2.5in 0.1in 2.5in, clip, scale=0.16]{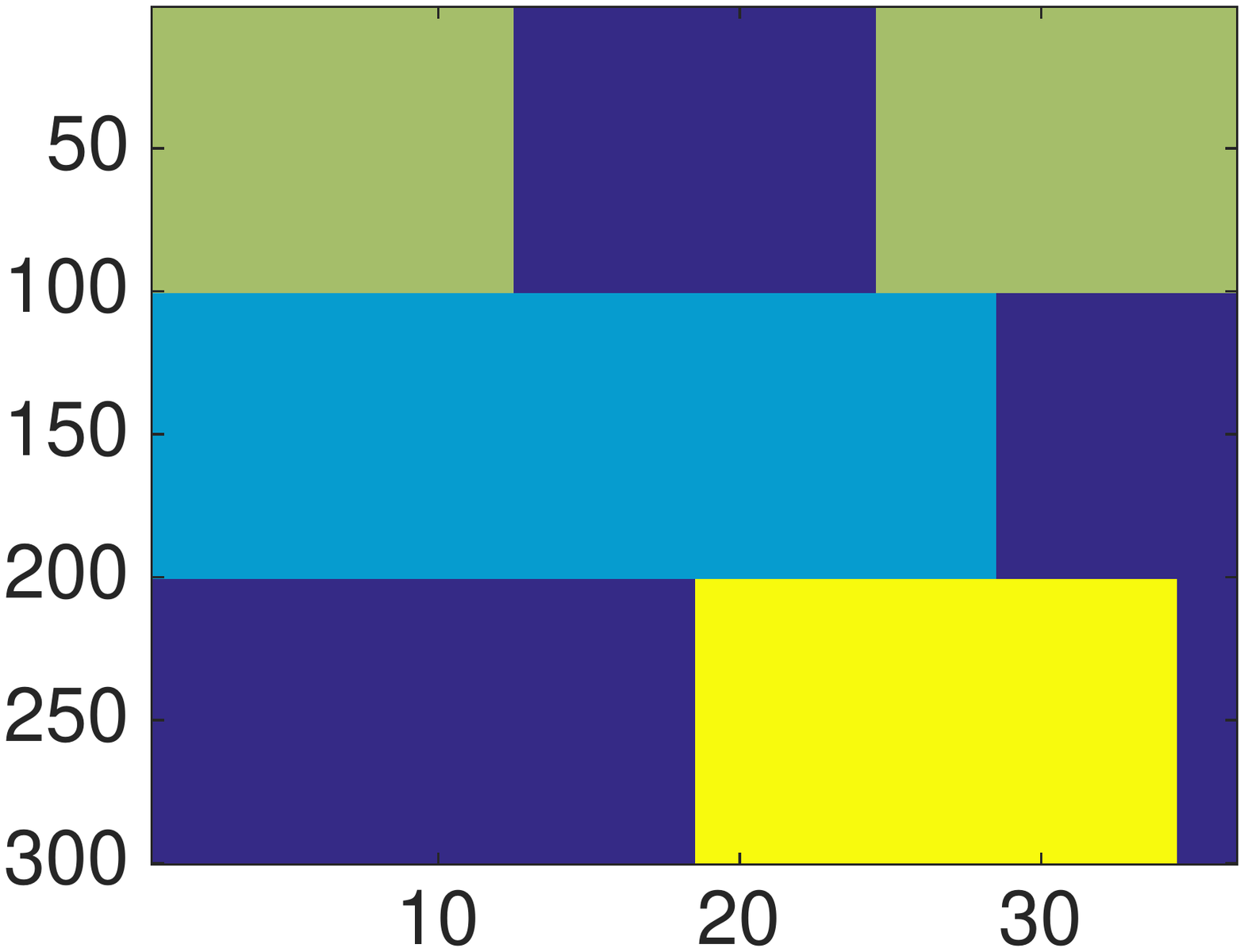}
   \label{subspace3d}
 }
 \label{subspace3}
\caption{\textbf{(Approximate budget)} CRAFT recovered the subspaces on both the categorical dataset shown in (a) and the numeric dataset shown in (c), and required minimal tuning.}    
\end{figure*}

We modified the original categorical synthetic dataset to have cluster 3 (a)  overlap with cluster 1,  and more importantly, (b) significantly overlap with cluster 2.  We obtained the configuration, shown in  Fig. \ref{subspace3a}, by splitting cluster 3 (8 features) evenly in two parts, and increasing the number of features in cluster 2 (16 features) considerably relative to cluster 1 (9 features), thereby making the distribution of features across the clusters non-uniform.  We observed, see Fig. \ref{subspace3b}, that for $\epsilon_c \in [0.76, 1)$, the CRAFT algorithm for the approximate budget setting recovered the subspace exactly for a wide range of $m$, more specifically for all values, when $m$ was varied in increments of 0.1 from $0.2$ to  $0.9$.  This implies the procedure essentially requires tuning only $\epsilon_c$. We easily found the appropriate range by searching in decrements of $0.01$ starting from 1.  Fig. \ref{subspace3d} shows the recovered subspaces for a similar set-up for the numeric data shown in Fig. \ref{subspace3c}. We observed that for $\epsilon_v \in [4, 6]$, the recovery  was robust to selection of $m \in [0.1, 0.9]$, similar to the case of categorical data.   For our purposes, we searched for $\epsilon_v$ in increments of 0.5 from 1 to 9, since the global variance was set to 9. Thus, with minimal tuning, we recovered subspaces in all cases.

\begin{figure*}[t]
\centering
\subfigure[$m$ = 0.2]{
   \includegraphics[trim = {0.3in 2.5in 0.1in 2.8in}, clip, scale=0.21]{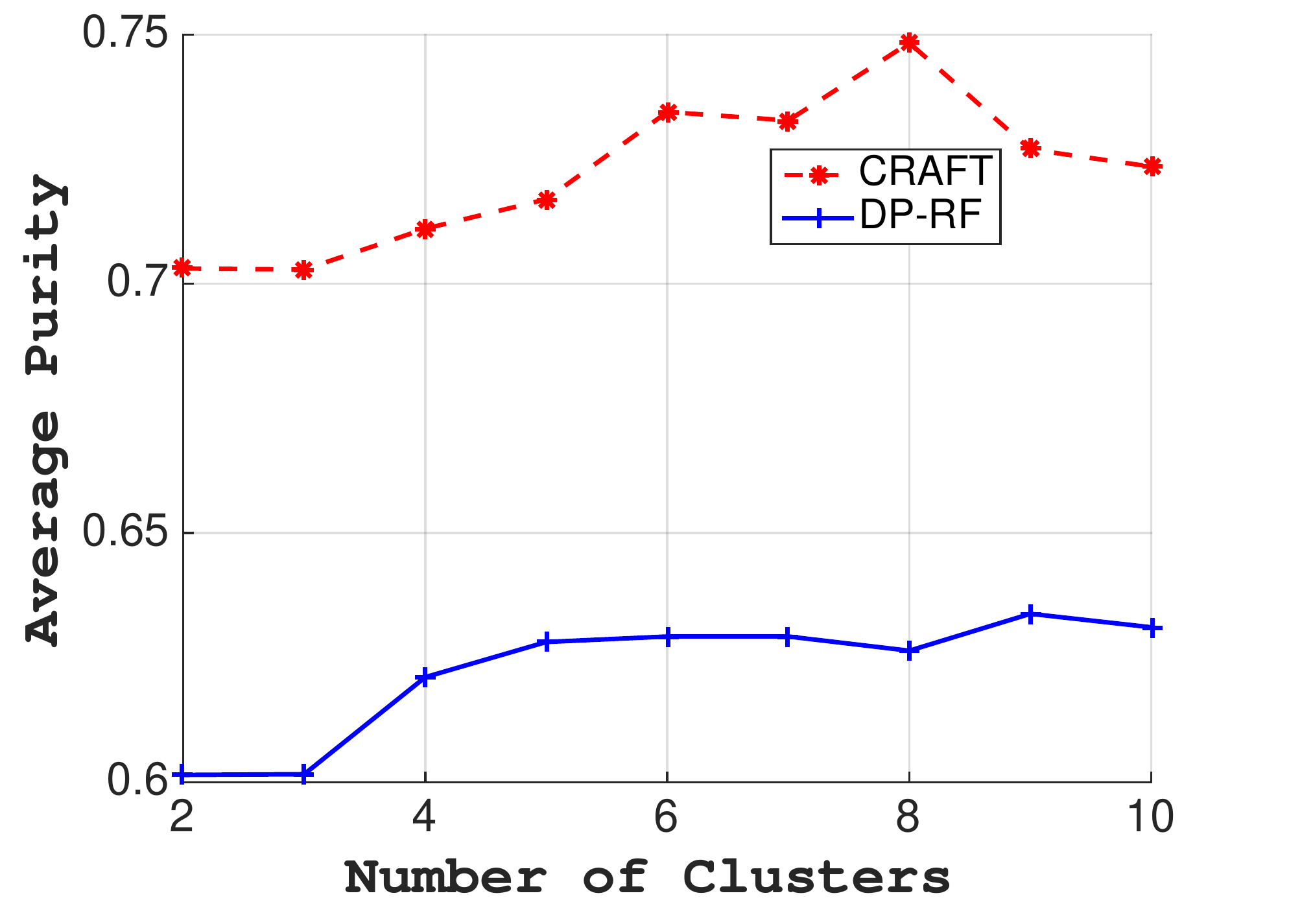}
  \label{Comparison1a}
 }
\subfigure[$m$ = 0.4]{
   \includegraphics[trim = {0.3in 2.5in 0.1in 2.8in}, clip, scale=0.21]{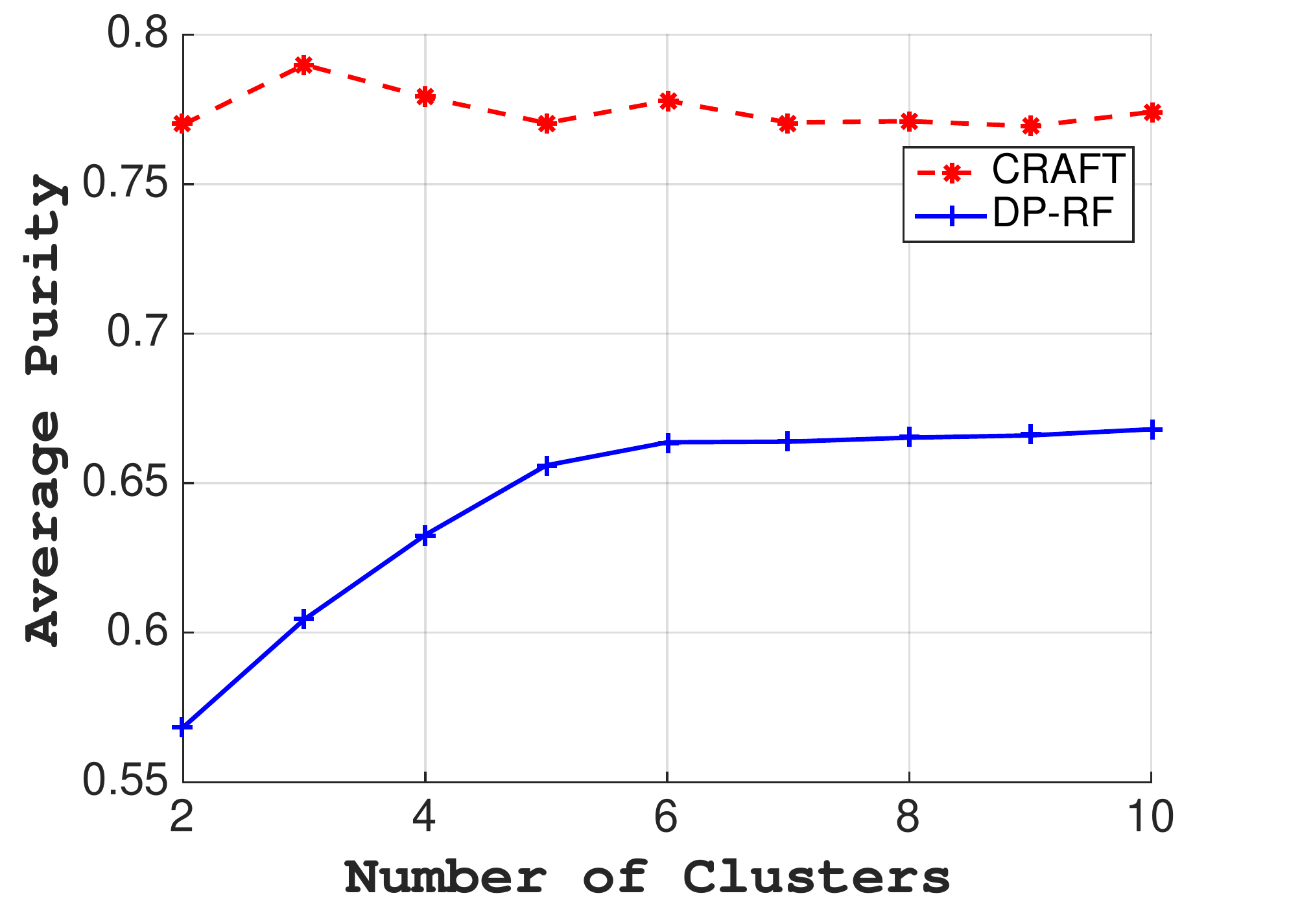}
   \label{Comparison1b}
   }
  \subfigure[$m$ = 0.6]{
   \includegraphics[trim = {0.3in 2.5in 0.1in 2.8in}, clip, scale=0.21]{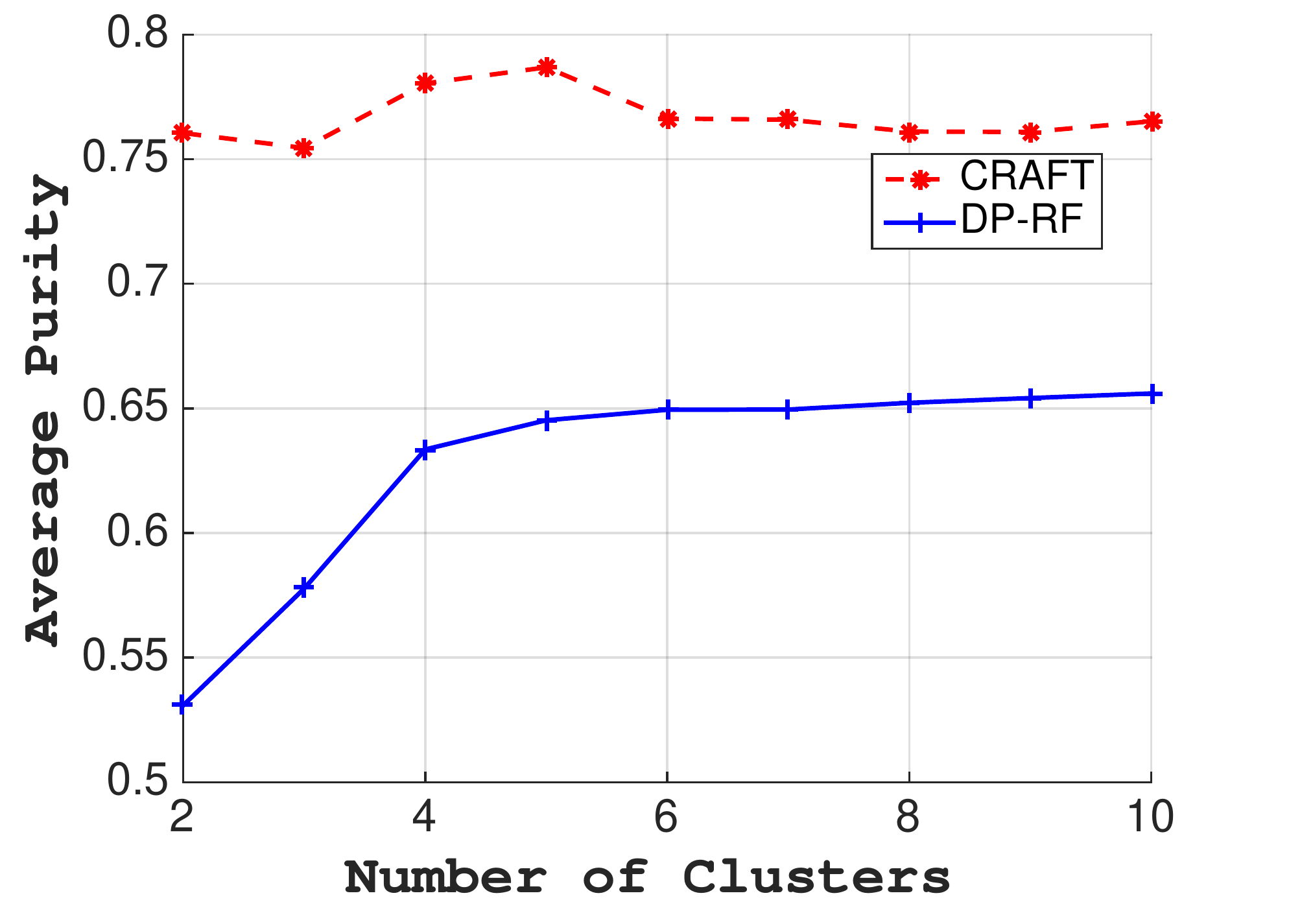}
  \label{Comparison1c}
 }
 \subfigure[$m$ = 0.2]{
   \includegraphics[trim = {0.3in 2.5in 0.1in 2.8in}, clip, scale=0.21]{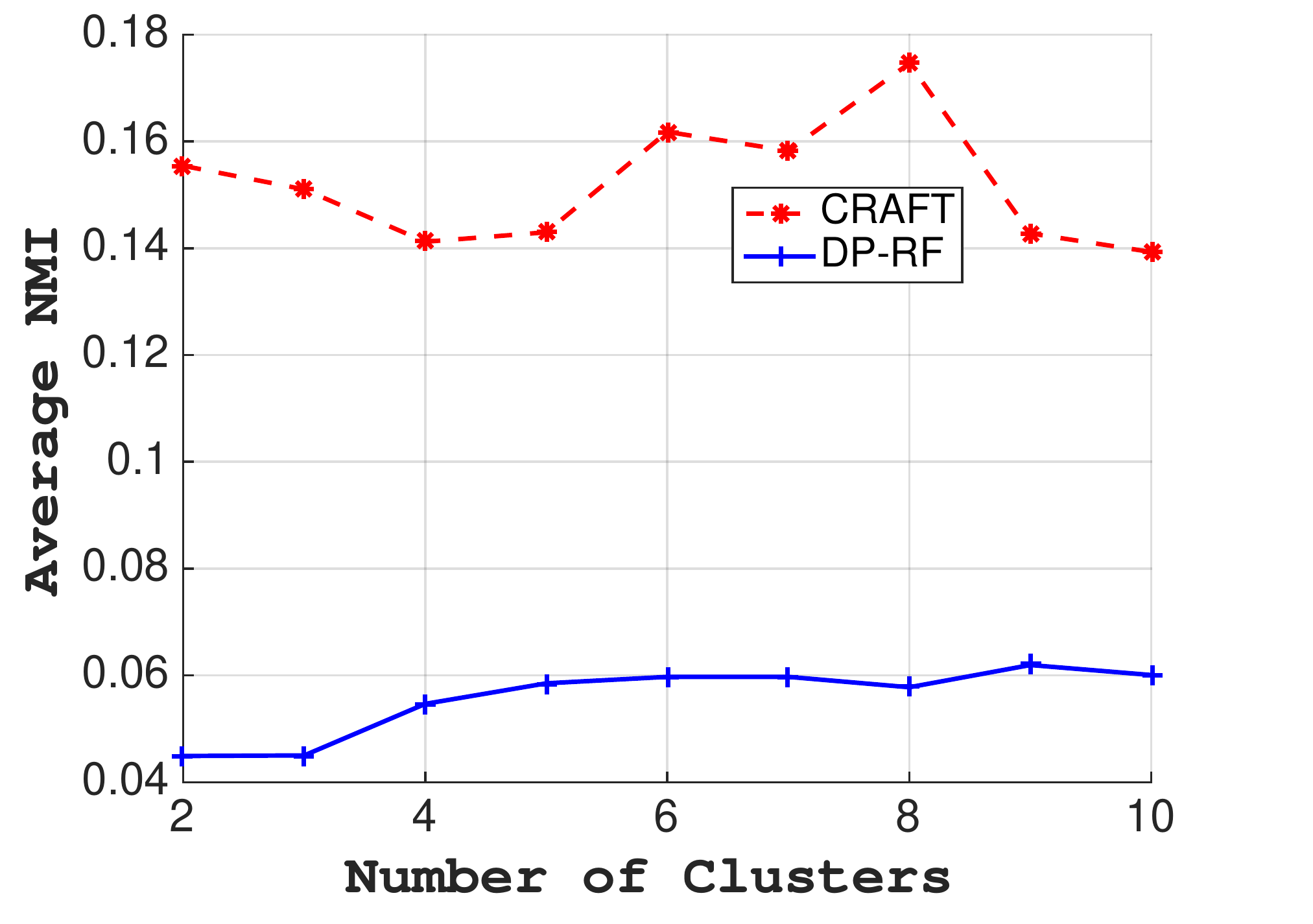}
  \label{Comparison2a}
 }
\subfigure[$m$ = 0.4]{
   \includegraphics[trim = {0.3in 2.5in 0.1in 2.8in}, clip, scale=0.21]{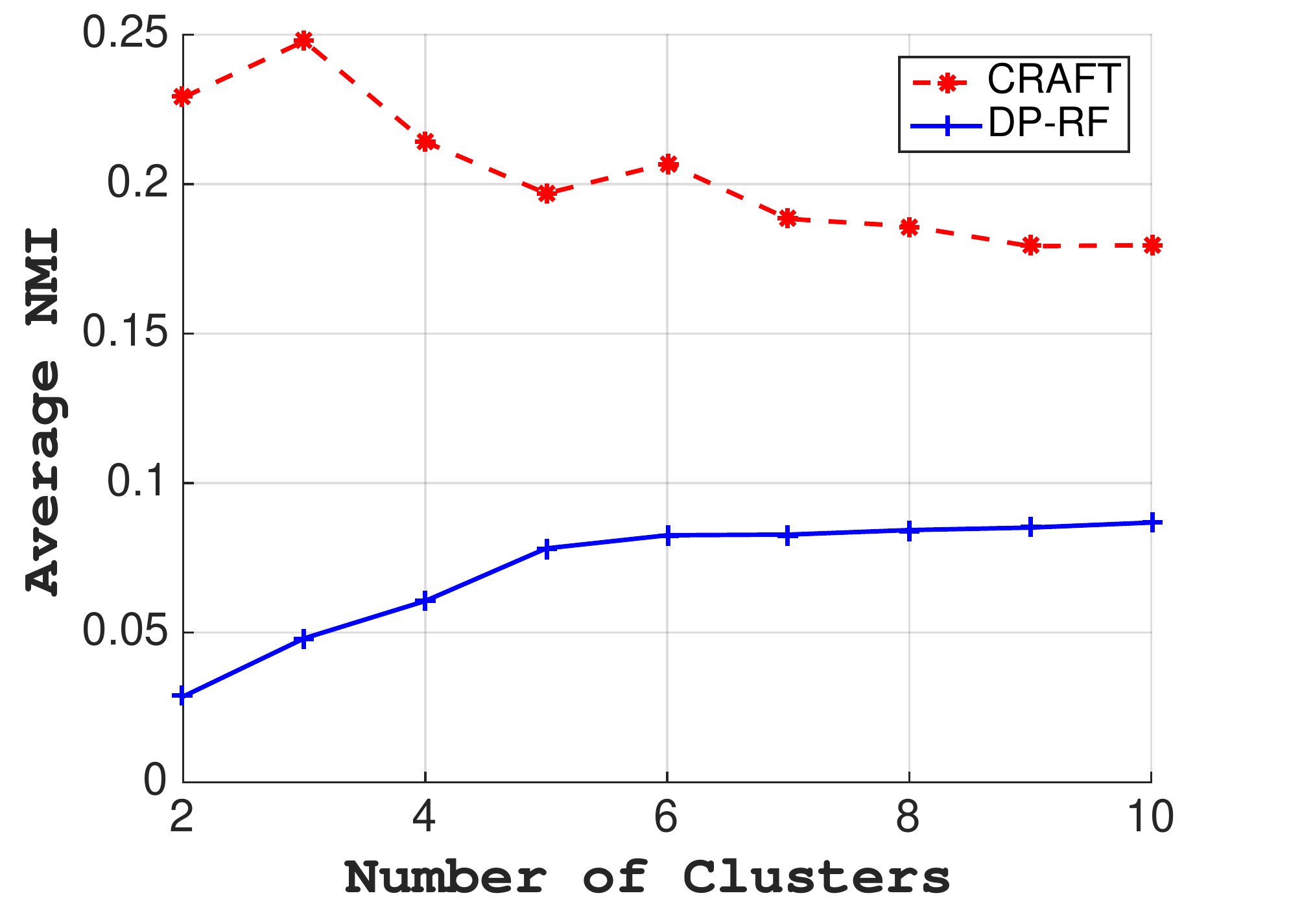}
   \label{Comparison2b}
 }
 \subfigure[$m$ = 0.6]{
   \includegraphics[trim = {0.3in 2.5in 0.1in 2.8in}, clip, scale=0.21]{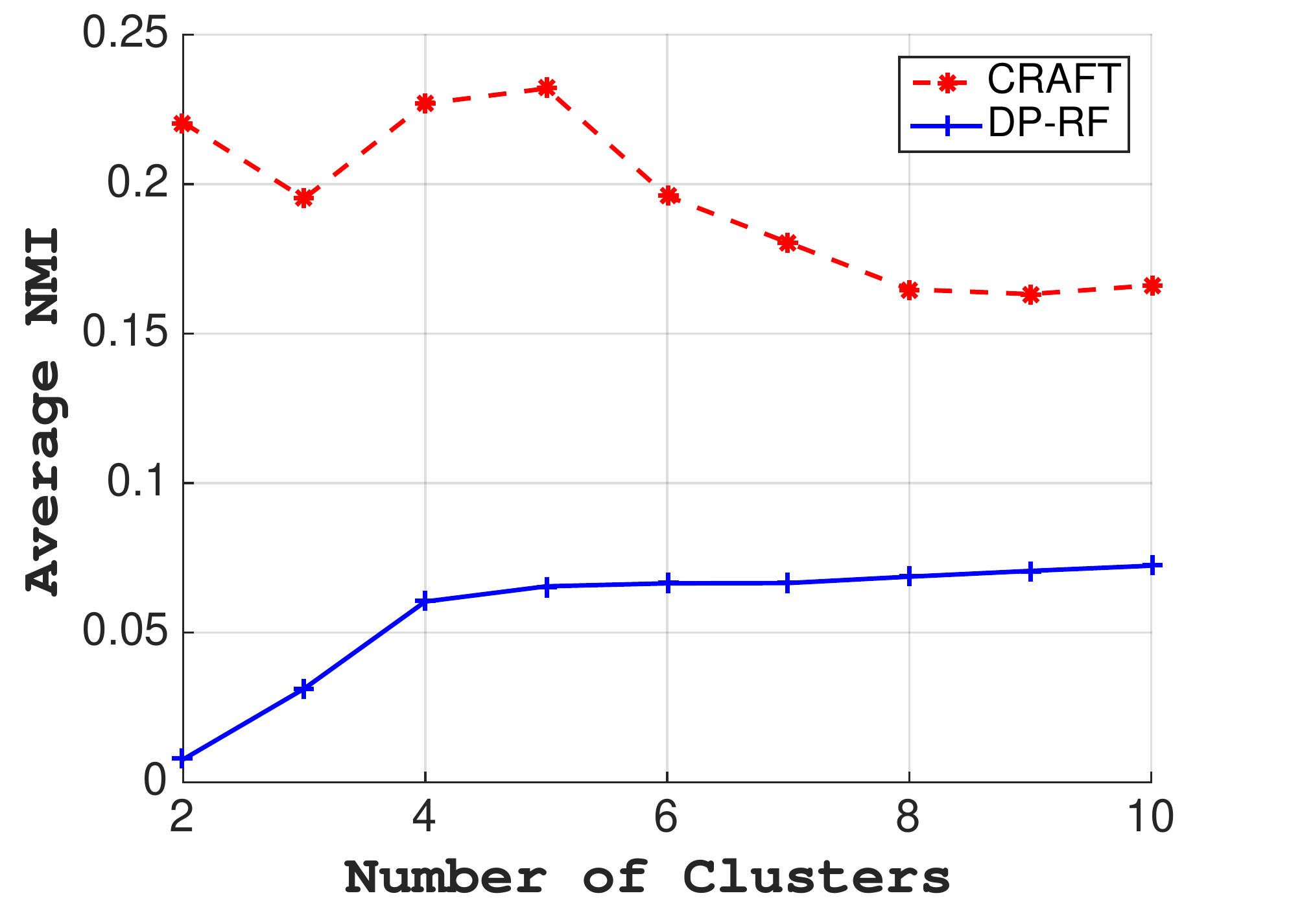}
  \label{Comparison2c}
  }
\caption{Purity (a-c) and NMI (d-f) comparisons on Splice for different values of $m$. DP-RF is DP-means(R) extended to incorporate feature selection.  \label{Comparison1}}
\end{figure*}

\subsubsection*{Experimental Setup for Real Datasets}
In order to compare the non-parametric CRAFT algorithm with other methods (where the number of clusters K is not defined in advance), we followed the farthest-first heuristic used by the authors of DP-means \cite{Kulis}, which is reminiscent of the seeding proposed in methods such as K-means$++$ \cite{KMeans++} and Hochbaum-Shmoys initialization \cite{HS}: for an approximate number of desired clusters $k$,  a suitable $\lambda$ is found in the following manner. First a singleton set $T$ is initialized, and then iteratively at each of the $k$ rounds, the point in the dataset that is farthest from $T$ is added to $T$. The distance of a point $x$ from $T$ is taken to be the smallest distance between $x$  and any point in $T$, for evaluating the corresponding objective function. At the end of the $k$ rounds, we set $\lambda$ as the distance of the last point that was included in $T$. Thus, for both DP-means and CRAFT, we determined their respective $\lambda$ by following the farthest first heuristic evaluated on their objectives: K-means objective for DP-means and entropy based objective for CRAFT.  

\par \citet{Kulis} initialized $T$ to the global mean for DP-means algorithm. We instead chose a point randomly from the input to initialize $T$ for CRAFT.  In our experiments, we found that this strategy can be often more effective than using the global mean since the cluster centers tend to be better separated and less constrained.  However, to highlight that the poor performance of DP-means is not just an artifact of the initial cluster selection strategy but more importantly, it is due to the mismatch of the Euclidean distance to categorical data,
 we also conducted experiments on DP-means with random selection of the initial cluster center from the data points. We call this method DP-means(R) where R indicates randomness in selecting the initial center. 

\subsubsection*{Evaluation Criteria For Real Datasets}
To evaluate the quality of clustering, we use datasets with known true labels. We use two standard metrics, \textit{purity} and \textit{normalized mutual information} (NMI), to measure the clustering performance \cite{ManningRaSc08, StrehlGh02}. To compute purity, each full cluster is assigned to the class label that is most frequent in the cluster. Purity is the proportion of examples that we assigned to the correct label. Normalized mutual information is the mutual information between the cluster labeling and the true labels, divided by the square root of the true label entropy times the clustering assignment entropy. Both purity and NMI lie between 0 and 1 -- the closer they are to 1, the better the quality of the clustering. 

Henceforth, we use Algorithm \ref{alg3} with the fixed budget setting in our experiments to ensure a fair comparison with the other methods, since they presume a fixed $m$. 

\begin{table*}[]
\centering
\caption{CRAFT versus DP-means and state-of-the-art feature selection methods when half of the features were selected (i.e. $m = 0.5$). We abbreviate MCFS to M, NDFS to N, DP-means to D, and DP-means(R) to DR to fit the table within margins. DP-means and  DP-means(R) do not select any features. The number of clusters was chosen to be same as the number of classes in each dataset. \vspace*{5pt}
\label{Table:SpliceAll}}
\begin{tabular}{|c|c|c|c|c|c||c|c|c|c|c|} \hline
\hline
 \textbf{Dataset} &\multicolumn{5}{|c||}{\textbf{Average Purity}}&\multicolumn{5}{|c|}{\textbf{Average NMI}}\\
\cline{2-11} 
 & \textbf{CRAFT} & \textbf{M} & \textbf{N} & \textbf{DR} & \textbf{D} & \textbf{CRAFT} & \textbf{M} & \textbf{N} & \textbf{DR} & \textbf{D}\\
\hline\hline
Bank   &   0.67  &  0.65 &    0.59 &   0.61&   0.61  &  0.16 &   0.06 &    0.02 &    0.03 &   0.03 \\ \hline
Spam    &        0.72 &   0.64 &    0.64 &   0.61 &    0.61 &  0.20 &    0.05 &   0.05 &   0.00 &      0.00\\ \hline
Splice    &       0.75  &  0.62  &  0.63   &  0.61    &   0.52  &     0.20  &   0.04  &  0.05  &  0.05    &   0.01 \\ \hline
Wine      &       0.71  &  0.72 &    0.69 &    0.66   &       0.66 &  0.47 &    0.35  &  0.47  &  0.44   &    0.44 \\ \hline
Monk     &       0.56  &  0.55 &   0.53  &  0.54   &    0.53 &  0.03 &    0.02  &  0.00  &  0.00   &    0.00 \\ \hline
\end{tabular}
\end{table*}

\begin{table*}[t!]
\centering
\caption{CRAFT versus DP-means and state-of-the-art feature selection methods ($m = 0.8$). \label{Table:SpliceAll2}}
\begin{tabular}{|c|c|c|c|c|c||c|c|c|c|c|} \hline
\hline
 \textbf{Dataset} &\multicolumn{5}{|c||}{\textbf{Average Purity}}&\multicolumn{5}{|c|}{\textbf{Average NMI}}\\
\cline{2-11} 
 & \textbf{CRAFT} & \textbf{M} & \textbf{N} & \textbf{DR} & \textbf{D} & \textbf{CRAFT} & \textbf{M} & \textbf{N} & \textbf{DR} & \textbf{D}\\
\hline\hline
Bank   &   0.64  &  0.61 &    0.61 &   0.61&   0.61  &  0.08 &   0.03 &    0.03 &    0.03 &   0.03 \\ \hline
Spam    &        0.72 &   0.64 &    0.64 &   0.61 &    0.61 &  0.23 &    0.05 &   0.05 &   0.00 &      0.00\\ \hline
Splice    &       0.74  &  0.68  &  0.63   &  0.61    &   0.52  &     0.18  &   0.09  &  0.05  &  0.05    &   0.01 \\ \hline
Wine      &       0.82  &  0.73 &    0.69 &    0.66   &       0.66 &  0.54 &    0.42  &  0.42  &  0.44   &    0.44 \\ \hline
Monk     &       0.57  &  0.54 &   0.54  &  0.54   &    0.53 &  0.03 &    0.00  &  0.00  &  0.00   &    0.00 \\ \hline
\end{tabular}
\end{table*}

\subsubsection*{Comparison of CRAFT with DP-means extended to Feature Selection} 
 We now provide evidence  that CRAFT outperforms DP-means on categorical data.   We use the Splice junction determination dataset \cite{Datasets} that has all categorical features. We borrowed the feature selection term from CRAFT to extend DP-means(R) to include feature selection, and retained its squared Euclidean distance measure. Recall that, in a special case, the CRAFT objective degenerates to DP-means(R) on numeric data when all features are retained, and cluster variances are all the same (see the Supplementary).  Fig. \ref{Comparison1}  shows the comparison results on the Splice data for different values of $m$. CRAFT outperforms extended DP-means(R) in terms of both purity and NMI, showing the importance of the entropy term in the context of clustering with feature selection.


\subsubsection*{Comparison with State-of-the-Art Unsupervised Feature Selection Methods}
We now demonstrate the benefits of cluster specific  feature selection accomplished by CRAFT.   Table \ref{Table:SpliceAll} and Table \ref{Table:SpliceAll2} show how CRAFT compares with two state-of-the-art unsupervised feature selection methods --  MCFS \cite{MCFS} and NDFS \cite{NDFS} -- besides DP-means and DP-means(R) on several datasets \cite{Datasets}, namely Bank, Spam, Wine,  Splice (described above),  and Monk, when $m$ was set to 0.5 and 0.8 respectively. Our experiments clearly highlight that CRAFT (a) works well for both numeric and categorical data, and (b) compares favorably with both the global feature selection algorithms and clustering methods, such as DP-means, that do not select features. 

Finally, we found that besides performance,  CRAFT also showed good performance in terms of time. For instance, on the Spam dataset for $m=0.5$, CRAFT required an average execution time of only 0.39 seconds, compared to 1.78 and 61.41 seconds by MCFS and NDFS respectively.  This behavior can be attributed primarily to the benefits of the scalable K-means style algorithm employed by CRAFT, as opposed to MCFS and NDFS that require computation-intensive spectral algorithms.   

\subsection*{Conclusion} CRAFT's framework incorporates cluster-specific feature selection and handles both categorical and numeric data. It can be extended in several ways, some of which are discussed in Section \ref{discussionsection}.  The objective obtained from MAP asymptotics is interpretable, and informs simple algorithms for both the fixed budget setting (the number of features selected per cluster is fixed) and the approximate budget setting (the number of features selected per cluster is allowed to vary across the clusters).  Code for CRAFT is available at the following website: \texttt{http://www.placeholder.com}.

\clearpage
\bibliographystyle{plainnat}
\bibliography{nips}

\cleardoublepage

\title{CRAFT: ClusteR-specific Assorted Feature selecTion (Supplementary)}
\maketitle
\section{Supplementary Material} \label{Supplementary}
We now derive the various objectives for the CRAFT framework. We first show the derivation for the generic objective that accomplishes feature selection on the assorted data. We then derive the degenerate cases when all features are retained and all data are (a)  numeric, and (b) binary categorical.  In particular, when the data are all numeric, we recover the DP-means objective \cite{Kulis}.  


\subsection{Main Derivation: Clustering with Assorted Feature Selection}
We have the total number of features, $D = |Cat| + |Num|$. We define $S_{N, k}$ to be the number of points assigned to cluster $k$. First, note that a Beta distribution with mean $c_1$ and variance $c_2$ has shape parameters  $\dfrac{c_1^2(1-c_1)}{c_2} - c_1$ and $\dfrac{c_1(1-c_1)^2}{c_2} + c_1 - 1$. Therefore, we can find the shape parameters corresponding to $m$ and $\rho$. Now, recall that for numeric data, we assume the density is of the following form:
 \begin{equation} \label{NumericDensity}
f(x_{nd}|v_{kd}) = \dfrac{1}{Z_{kd}} e^{-\left[v_{kd} \dfrac{(x_{nd}-\zeta_{kd})^2}{2 \sigma_{kd}^2} + (1-v_{kd}) \dfrac{(x_{nd}-\zeta_{d})^2}{2 \sigma_d^2}\right]},
 \end{equation}
 where $Z_{kd}$ ensures that the area under the density is 1.  Assuming an uninformative conjugate prior on the (numeric) means, i.e. a Gaussian distribution with infinite variance,  and using the Iverson bracket notation for discrete (categorical) data, we obtain the joint distribution given in Fig. \ref{bigeqns_general} for the underlying graphical model shown in Fig. \ref{fig:FDPCRAFT_mix-Graphical}.

\begin{figure} [h]    
\begin{eqnarray}
\lefteqn{\mathbb{P}(x, z, v, \nu, \eta, \zeta, m)}\nonumber\\ & = & \mathbb{P}(x|z, v, \eta, \zeta) \mathbb{P} (v|\nu) \mathbb{P}(z) \mathbb{P}(\eta) \mathbb{P}(\nu; m, \rho)  \nonumber\\
& = & \prod_{k=1}^{K^+} \prod_{n: z_{n, k} = 1} \Bigg[\left(\prod_{d \in Cat: v_{kd} = 1} \prod_{t \in \mathcal{T}_d} \eta_{kdt}^{\mathbb{I}(x_{nd} = t)}\right) \left(\prod_{d \in Cat: v_{kd} = 0} \prod_{t \in \mathcal{T}_d} \eta_{0dt}^{\mathbb{I}(x_{nd} = t)}\right) \nonumber \\ 
 &&  \left(\prod_{d' \in Num} \dfrac{1}{Z_{kd'}} e^{-\left[v_{kd'}(x_{nd'}-\zeta_{kd'})^2/(2 \sigma_{kd'}^2) + (1-v_{kd'}) (x_{nd'}-\zeta_{d'})^2/(2 \sigma_{d'}^2))\right]} \right) \Bigg] \nonumber \\
& & \cdot  \left[\prod_{k=1}^{K^+} \prod_{d=1}^D \nu_{kd}^{v_{kd}} (1 - \nu_{kd})^{1 - v_{kd}}\right] \cdot \left[\theta^{K^+ - 1} \dfrac{\Gamma\left(\theta+1\right)}{\Gamma\left(\theta + N\right)} \prod_{k=1}^{K^+} (S_{N, k} - 1)!\right]  \\
&&  \cdot \left[\prod_{k=1}^{K^+} \prod_{d \in Cat}  \dfrac{\Gamma\left( \sum_{t \in \mathcal{T}_d} \dfrac{\alpha_{kdt}}{K^+}\right)}{ \prod_{t \in \mathcal{T}_d} \Gamma\left(\dfrac{\alpha_{kdt}}{K^+}\right)}   \prod_{t' \in \mathcal{T}_d} \eta_{kdt'}^{(\alpha_{kdt'}/K^+)-1}\right]\nonumber \\ 
& & \cdot \prod_{k=1}^{K^+}  \prod_{d=1}^D \dfrac{ \Gamma \left(\dfrac{m(1-m)}{\rho} - 1\right) \nu_{kd}^{\left(\dfrac{m^2(1-m)}{\rho} - m - 1\right)} (1-\nu_{kd})^{\left(\dfrac{m(1-m)^2}{\rho} - (2-m) \right)} }{\Gamma \left(\dfrac{m^2(1-m)}{\rho} - m\right) \Gamma \left(\dfrac{m(1-m)^2}{\rho} - (1-m) \right) }  \nonumber
\end{eqnarray}
\caption{Joint probability distribution for the generic case (both numeric and categorical features).\label{bigeqns_general}}
\end{figure}

The total contribution of \eqref{NumericDensity} to the negative joint log-likelihood 
\begin{eqnarray}  \label{Contribution1}
 =   \sum_{k=1}^{K^+} \sum_{d \in Num} \sum_{n: z_{n, k} = 1} \bigg[ v_{kd} \dfrac{(x_{nd}-\zeta_{kd})^2}{2 \sigma_{kd}^2} + (1-v_{kd}) \dfrac{(x_{nd}-\zeta_{d})^2}{2 \sigma_d^2} \bigg] + \sum_{k=1}^{K^+} \sum_{d \in Num}  \log\, Z_{kd}.
\end{eqnarray}

The contribution of the selected categorical features depends on the categorical means of the clusters, and is given by
\begin{eqnarray*}
 - \log\, \left(\prod_{k=1}^{K^+} \prod_{n: z_{n, k} = 1} \prod_{d \in Cat: v_{kd} = 1} \prod_{t \in \mathcal{T}_d} \eta_{kdt}^{\mathbb{I}(x_{nd} = t)}\right). 
 \end{eqnarray*}
 On the other hand, the categorical features not selected are assumed to be drawn from cluster-independent global means, and therefore contribute
 \begin{eqnarray*}
 - \log\, \left(\prod_{k=1}^{K^+} \prod_{n: z_{n, k} = 1} \prod_{d \in Cat: v_{kd} = 0} \prod_{t \in \mathcal{T}_d} \eta_{0dt}^{\mathbb{I}(x_{nd} = t)} \right).\\ 
 \end{eqnarray*}
Thus, the total contribution of the categorical features is
\begin{eqnarray} \label{Contribution2} - \sum_{k=1}^{K^+} \sum_{n: z_{n, k} = 1} \left[\sum_{d \in Cat: v_{kd} = 1} \sum_{t \in \mathcal{T}_d} \mathbb{I}(x_{nd} = t) \log\, \eta_{kdt}
 +  \sum_{d \in Cat: v_{kd} = 0} \sum_{t \in \mathcal{T}_d} \mathbb{I}(x_{nd} = t) \log\, \eta_{0dt}\right].  \nonumber
 \end{eqnarray}

The Bernoulli likelihood on $v_{kd}$ couples with the conjugate Beta prior on $\nu_{kd}$. To avoid having to provide the value of $\nu_{kd}$ as a parameter, we take its point estimate to be the mean of the resulting Beta posterior, i.e., we set 
\begin{equation} \label{nu} 
 \nu_{kd} = \dfrac{\left(\dfrac{m^2(1-m)}{\rho} - m\right) +  v_{kd}}{\dfrac{m(1-m)}{\rho} } = \dfrac{a_{kd}}{a_{kd}+b_{kd}},
 \end{equation}
where 
\begin{eqnarray}
a_{kd} & = & \dfrac{m^2(1-m)}{\rho} - m +  v_{kd}, \text{ and} \nonumber \\
b_{kd} & = & \dfrac{m(1-m)^2}{\rho} + m -  v_{kd} \label{eqNu}. \nonumber
\end{eqnarray}
Then the contribution of the posterior to the negative log likelihood is
$$ - \sum_{k=1}^{K^+} \sum_{d= 1}^{D} \bigg[\log \, \left(\dfrac{a_{kd}}{a_{kd} + b_{kd}}\right)^{a_{kd}}  +   \log \, \left(\dfrac{b_{kd}}{a_{kd} + b_{kd}}\right)^{b_{kd}}  \bigg], $$
or equivalently,
$$\sum_{k=1}^{K^+} \sum_{d= 1}^{D} \underbrace{\left[\log\, (a_{kd} + b_{kd})^{(a_{kd} + b_{kd})} - \log \, a_{kd}^{a_{kd}} - \log \, b_{kd}^{b_{kd}}  \right]}_{F(v_{kd})}.$$
Since $v_{kd} \in \{0, 1\}$, this simplifies to
\begin{eqnarray} \label{Contribution3}
\displaystyle \sum_{k=1}^{K^+} \sum_{d= 1}^{D} F(v_{kd})  =  \displaystyle \sum_{k=1}^{K^+} \sum_{d= 1}^{D}  \left[v_{kd}(F(1) - F(0)) + F(0)\right] =  \left(\displaystyle \sum_{k=1}^{K^+} \sum_{d= 1}^{D} v_{kd}\right) \Delta F + K^{+} DF(0),
\end{eqnarray}
where $\Delta F = F(1) - F(0)$ quantifies the change when a feature is selected for a cluster.

The numeric means do not make any contribution since we assumed an uninformative conjugate prior over $\mathbb{R}$.  On the other hand, the categorical means contribute
$$- \log\, \left[\prod_{k=1}^{K^+} \prod_{d \in Cat}  \dfrac{\Gamma\left( \sum_{t \in \mathcal{T}_d} \dfrac{\alpha_{kdt}}{K^+}\right)}{ \prod_{t \in \mathcal{T}_d} \Gamma\left(\dfrac{\alpha_{kdt}}{K^+}\right)}   \prod_{t' \in \mathcal{T}_d} \eta_{kdt'}^{(\alpha_{kdt'}/K^+)-1}\right],$$
which simplifies to
\begin{eqnarray} \label{Contribution4}
\sum_{k=1}^{K^+} \sum_{d \in Cat}  \left[-\log\, \dfrac{\Gamma\left( \sum_{t \in \mathcal{T}_d} \dfrac{\alpha_{kdt}}{K^+}\right)}{ \prod_{t \in \mathcal{T}_d} \Gamma\left(\dfrac{\alpha_{kdt}}{K^+}\right)} -   \sum_{t' \in \mathcal{T}_d} \left(\dfrac{\alpha_{kdt'}}{K^+} - 1\right) \log\,  \eta_{kdt'}\right].
\end{eqnarray}

Finally, the Dirichlet process specifies a distribution over possible clusterings, while favoring assignments of points to a small number of clusters. The contribution of the corresponding term is 
 $$- \log\, \left[\theta^{K^+ - 1} \dfrac{\Gamma\left(\theta+1\right)}{\Gamma\left(\theta + N\right)} \prod_{k=1}^{K^+} (S_{N, k} - 1)!\right],$$
 or equivalently,
\begin{eqnarray} \label{Contribution5}
  -(K^+ - 1) \log \theta - \log\, \left(\dfrac{\Gamma\left(\theta+1\right)}{\Gamma\left(\theta + N \right)} \prod_{k=1}^{K^+} (S_{N, k} - 1)!\right). 
\end{eqnarray}
The total negative log-likelihood is just the sum of terms in \eqref{Contribution1}, \eqref{Contribution2}, \eqref{Contribution3}, \eqref{Contribution4}, and \eqref{Contribution5}. We want to maximize the joint likelihood, or equivalently, minimize the total negative log-likelihood.  We would use asymptotics to simplify our objective. In particular, letting $\sigma_{d} \to \infty, \,\,\forall k \in [K^+] \text{ and } d \in Num$,  and $\alpha_{kdt} \to K^{+},\,\,  \forall t \in \mathcal{T}_{d},\,\, d \in Cat, k \in [K^+]$, and setting $\log \, \theta$ to
 $$ -\left(\lambda + \dfrac{\displaystyle \sum_{k=1}^{K^+} \displaystyle \sum_{d \in Cat} \log\, |\mathcal{T}_{d}|   - \displaystyle \sum_{k=1}^{K^+} \displaystyle \sum_{d \in Num}  \log\, Z_{kd}}{K^+-1}\right),$$
 we obtain our objective for assorted feature selection:
 
\begin{eqnarray}\nonumber 
\arg\!\!\!\!\!\!\!\!\!\!\min_{z, v, \eta, \zeta, \sigma}\, \underbrace{\sum_{k=1}^{K^+} \sum_{n: z_{n, k} = 1} \sum_{d \in Cat} \left[-v_{kd} \sum_{t \in \mathcal{T}_d} \mathbb{I}(x_{nd} = t) \log \eta_{kdt} -  (1-v_{kd}) \sum_{t \in \mathcal{T}_d} \mathbb{I}(x_{nd} = t) \log \eta_{0dt} \right]}_{\text{Categorical Data Discrepancy}} \\
+  \underbrace{\sum_{k=1}^{K^+} \sum_{n: z_{n, k} = 1}  \sum_{d \in Num} v_{kd} \dfrac{(x_{nd} - \zeta_{kd})^2}{2\sigma_{kd}^2}}_{\text{Numeric Data Discrepancy}} + \underbrace{(\lambda + DF_0) K^+}_{\text{Regularization Term}} + \underbrace{\left(\displaystyle \sum_{k=1}^{K^+} \sum_{d= 1}^{D} v_{kd}\right) F_{\Delta}}_{\text{Feature Control}}, \nonumber
\end{eqnarray}    \normalsize
where $\Delta F = F(1) - F(0)$ quantifies the change when a feature is selected for a cluster, and we have renamed the constants $F(0)$ and $\Delta F$ as $F_0$ and $F_{\Delta}$ respectively. 
   
\subsubsection{Setting $\rho$ \label{SubsecGuide}}
Reproducing the equation for $\nu_{kd}$ from \eqref{nu},  since we want to ensure that $\nu_{kd} \in (0, 1)$, we must have
$$0 < \dfrac{\left(\dfrac{m^2(1-m)}{\rho} - m\right) +  v_{kd}}{\dfrac{m(1-m)}{\rho}} <1.$$
Since $v_{kd} \in \{0, 1\}$, this immediately constrains
$$\rho \in (0, m(1-m)).$$
Note that $\rho$ guides the selection of features: a high value of $\rho$, close to $m(1-m)$, enables local feature selection ($v_{kd}$ becomes important), whereas a low value of $\rho$, close to 0, reduces the influence of $v_{kd}$ considerably, thereby resulting in global selection.

\subsection{Degenerate Case: Clustering Binary Categorical Data without Feature Selection}
In this case,  the discrete distribution degenerates to Bernoulli, while the numeric discrepancy and the feature control terms do not arise. Therefore, we can replace the Iverson bracket notation by having cluster means $\mu$ drawn from Bernoulli distributions. Then, the joint distribution of the observed data $x$, cluster indicators $z$  and cluster means $\mu$ is given by 
\begin{eqnarray}
\mathbb{P}(x, z, \mu) &=& \mathbb{P}(x|z, \mu) \mathbb{P}(z) \mathbb{P}(\mu) \nonumber\\
&=& \underbrace{\left[\prod_{k=1}^{K^+} \prod_{n: z_{n, k} = 1} \prod_{d=1}^D \mu_{kd}^{x_{nd}} (1-\mu_{kd})^{1-x_{nd}}\right]}_{(A)}  \cdot \underbrace{\left[\theta^{K^+ - 1} \dfrac{\Gamma\left(\theta+1\right)}{\Gamma\left(\theta + N\right)} \prod_{k=1}^{K^+} (S_{N, k} - 1)!\right]}_{(B)} \nonumber \\
&& \cdot \underbrace{\left[\prod_{k=1}^{K^+} \prod_{d=1}^D \dfrac{\Gamma\left(\dfrac{\alpha}{K^+} + 1\right)}{\Gamma\left(\dfrac{\alpha}{K^+}\right) \Gamma(1)} \mu_{kd}^{\frac{\alpha}{K^+} -1} (1-\mu_{kd})^0\right]}_{(C)}. \nonumber 
\end{eqnarray} 
The joint negative log-likelihood is
\begin{equation*}
 - \log \mathbb{P}(x, z, \mu)  =   -[\log\, (A) + \log\, (B) + \log\, (C)]. 
\end{equation*}
We first note that
\begin{eqnarray*}
\log\, (A) &= &\sum_{k=1}^{K^+} \sum_{n: z_{n, k} = 1} \sum_{d=1}^D x_{nd} \log \mu_{kd} + (1-x_{nd}) \log (1-\mu_{kd}) \nonumber \\
& = & \sum_{k=1}^{K^+} \sum_{n: z_{n, k} = 1} \sum_{d=1}^D x_{nd} \log\left(\dfrac{\mu_{kd}}{1-\mu_{kd}} \right) + \log (1-\mu_{kd}) \nonumber \\
& = & \sum_{k=1}^{K^+} \sum_{n: z_{n, k} = 1} \sum_{d=1}^D  \bigg[\log (1-\mu_{kd}) + \mu_{kd} \log\left( \dfrac{\mu_{kd}}{1-\mu_{kd}} \right) \nonumber\\ 
&& +\,\,   x_{nd} \log\left(\dfrac{\mu_{kd}}{1-\mu_{kd}} \right) \nonumber -  \mu_{kd} \log\, \left( \dfrac{\mu_{kd}}{1-\mu_{kd}} \right)\bigg]  \nonumber\\
& = & \sum_{k=1}^{K^+} \sum_{n: z_{n, k} = 1} \sum_{d=1}^D \bigg[(x_{nd} - \mu_{kd}) \log\left(\dfrac{\mu_{kd}}{1-\mu_{kd}} \right) \nonumber\\
&& +\,\, \mu_{kd} \log \mu_{kd} + (1 - \mu_{kd}) \log (1 - \mu_{kd})\bigg] \nonumber\\
& = & \sum_{k=1}^{K^+} \sum_{n: z_{n, k} = 1} \sum_{d=1}^D (x_{nd} - \mu_{kd}) \log\left(\dfrac{\mu_{kd}}{1-\mu_{kd}} \right) - \mathbb{H}(\mu_{kd}),  \nonumber
\end{eqnarray*}
where
\[\mathbb{H}(p)  =  -p \log p - (1 - p) \log (1 - p)\, \text{ for } p \in [0, 1].  \]

$\log\, (B)$ and  $\log\, (C)$ can be computed via steps analogous to those used in assorted feature selection. Invoking the asymptotics by letting $\alpha \to K^+$, and  setting $$\theta = e^{-\left(\lambda + \dfrac{K^+D}{K^+-1} \log \left(\dfrac{\alpha}{K^+} \right)\right)},$$ we obtain the following objective:
\begin{eqnarray} \label{BinaryClusterObjective1}
\arg\!\min_{z, \mu}\,\,  \sum_{k=1}^{K^+} \sum_{n: z_{n, k} = 1}  \sum_{d} \underbrace{\bigg[ \mathbb{H}(\mu_{kd}) \, + \,  (\mu_{kd} - x_{nd}) \log\left(\dfrac{\mu_{kd}}{1-\mu_{kd}} \right) \bigg]}_{(\text{Binary Discrepancy})} +  \lambda K^+, 
\end{eqnarray}  
where the term (Binary Discrepancy) is an objective for binary categorical data, similar to the K-means objective for numeric data. This suggests a very intuitive procedure, which is outlined in Algorithm \ref{alg2}.

\begin{algorithm}                      
\caption{Clustering binary categorical data}          
\label{alg2}                           
\begin{algorithmic}                    
    \REQUIRE $x_1, \ldots, x_N \in \{0, 1\}^{D}$: binary categorical data, and $\lambda > 0$: cluster penalty parameter.
    \ENSURE $K^+$: number of clusters and $l_1, \ldots, l_{K^+}$: clustering.
    \begin{enumerate} 
     \STATE Initialize $K^+ = 1$, $l_1 = \{x_1, \ldots, x_N\}$ and the mean $\mu_{1}$ (sample randomly from the dataset).
     \STATE Initialize cluster indicators $z_n = 1$ for all $n \in [N]$.
     \STATE Repeat until convergence
     \begin{itemize}
          \item Compute $\forall k \in [K^+], d \in [D]$: \begin{eqnarray*}\hspace*{-10pt}\mathbb{H}(\mu_{kd})  =  - \mu_{kd} \log \mu_{kd} - (1 - \mu_{kd}) \log (1 - \mu_{kd}).\end{eqnarray*}
           \item For each point $x_n$ 
            \begin{itemize}
                \item \hspace*{-0.2 cm} Compute the following for all $k\in [K^+]$:
                \begin{eqnarray*} \hspace*{-50pt}
                d_{nk} = \sum_{d=1}^D \left[\mathbb{H}(\mu_{kd}) + (\mu_{kd} - x_{nd}) \log\left(\dfrac{\mu_{kd}}{1-\mu_{kd}} \right) \right]. 
                \end{eqnarray*}
                 \STATE If $\displaystyle \min_k d_{nk} > \lambda$, set $K^+ = K^+ + 1$, $z_n = K^+$, and $\mu_{K^+} = x_n$.  
                 \STATE Otherwise, set $z_n = \displaystyle \arg\!\min_k d_{nk}$.
            \end{itemize}  
            \item Generate clusters $l_1, \ldots, l_{K^+}$ based on $z_1, \ldots, z_{K^+}$: $l_k = \{x_n \,|\, z_n = k\}$.
            \item For each cluster $l_k$, update $\mu_{k} = \dfrac{1}{|l_k|} \displaystyle \sum_{x \in l_k} x$. 
     \end{itemize}
    \end{enumerate}
\end{algorithmic}
\end{algorithm}

In each iteration, the algorithm computes ``distances" to the cluster means for each point to the existing cluster centers, and checks if the minimum distance is within $\lambda$. If yes, the point is assigned to the nearest cluster, otherwise a new cluster is started with the point as its cluster center. The cluster means are updated at the end of each iteration, and the steps are repeated until there is no change in cluster assignments over successive iterations. 

We get a more intuitively appealing objective by noting that the objective  \eqref{BinaryClusterObjective1} can be equivalently written as 
\small \begin{eqnarray}\label{BinaryClusterObjective2}
\arg\!\min_{z}\,\,  \sum_{k=1}^{K^+} \sum_{n: z_{n, k} = 1}  \sum_{d} \mathbb{H}(\mu_{kd}^*)  +\,\,  \lambda K^+,
\end{eqnarray} \normalsize
where $\mu_{kd}^*$ denotes the mean of feature $d$ computed by using points belonging to cluster $k$. characterizes the uncertainty. Thus the objective tries to minimize the overall uncertainty across clusters and thus forces similar points to come together.  
The regularization term ensures that the points do not form too many clusters, since in the absence of the regularizer each point will form a singleton cluster thereby leading to a trivial clustering. 

%
%

\subsection{Degenerate Case: Clustering Numerical Data without Feature Selection}
In this case, there are no categorical terms. Furthermore,  assuming an uninformative conjugate prior on the numeric means,  the terms that contribute to the negative joint log-likelihood are 
$$\prod_{k=1}^{K^+} \prod_{d'} \dfrac{1}{Z_{kd'}} e^{-\left[v_{kd'}(x_{nd'}-\zeta_{kd'})^2/(2\sigma_{kd'}^2) + (1-v_{kd'}) (x_{nd'}-\zeta_{d'})^2/(2\sigma_{d'}^2)\right]},$$
and 
$$\theta^{K^+ - 1} \dfrac{\Gamma\left(\theta+1\right)}{\Gamma\left(\theta + N\right)} \prod_{k=1}^{K^+} (S_{N, k} - 1)!.$$
Taking the negative logarithms on both these terms and adding them up, setting $\log \, \theta$ to  
$$-\left(\lambda + \dfrac{\displaystyle \sum_{k=1}^{K^+} \displaystyle \sum_{d'}  \log\, Z_{kd'}}{K^+-1}\right),$$
and $v_{kd'} = 1$ (since all features are retained), and letting $\sigma_{d'} \to \infty$ for all $d'$, we obtain

 \begin{eqnarray} \arg\!\min_{z}\,\, \sum_{k=1}^{K^+} \sum_{n: z_{n, k} = 1}  \sum_{d} \dfrac{(x_{nd} - \zeta^*_{kd})^{2}}{2 \sigma_{kd}^{*2}} +  \lambda K^+, \end{eqnarray}
 where $\zeta^*_{kd}$ and $\sigma_{kd}^{*2}$ are, respectively, the mean and variance of the feature $d$ computed using all the points assigned to cluster $k$. This degenerates to the DP-means objective \cite{Kulis} when $\sigma^*_{kd} = 1/\sqrt{2}$, for all $k$ and $d$. Thus, using a completely different model and analysis to \cite{Kulis}, we recover the DP-means objective as a special case. 
\cleardoublepage

\end{document}